\documentclass{article}

\usepackage[nonatbib, final]{neurips_2021}

\usepackage[utf8]{inputenc} % allow utf-8 input
\usepackage[T1]{fontenc}    % use 8-bit T1 fonts
\usepackage[dvipsnames]{xcolor}
\usepackage[pagebackref=true,breaklinks=true,letterpaper=true,colorlinks,bookmarks=false, citecolor=ForestGreen]{hyperref}     % hyperlinks
\usepackage{url}            % simple URL typesetting
\usepackage{booktabs}       % professional-quality tables
\usepackage{amsfonts}       % blackboard math symbols
\usepackage{nicefrac}       % compact symbols for 1/2, etc.
\usepackage{microtype}      % microtypography

\usepackage{graphicx}
\usepackage{comment}
\usepackage{amsmath,amssymb} % define this before the line numbering.
\usepackage{xspace}
\usepackage[font=footnotesize,labelfont=bf]{caption} 
\usepackage{caption}
\usepackage{mathtools}
\usepackage{graphicx}
\usepackage{soul}
\usepackage{xspace}
\usepackage{cleveref}
\usepackage{enumitem}
\usepackage[square,numbers,sort&compress]{natbib}

\usepackage{bbding}
\usepackage{wasysym}
\usepackage{amssymb}
\usepackage{float}
\usepackage{dblfloatfix}
\usepackage{footnote}
\usepackage{array} % For tables
\usepackage{dblfloatfix}
\usepackage{transparent}
\usepackage{xspace}
\usepackage{multirow}
\usepackage{amsfonts}
\usepackage{pifont}
\usepackage{algorithm}
\usepackage{listings}
\usepackage{bbold}
\usepackage{wrapfig}
\usepackage{subcaption}

\usepackage{etoolbox} % from moco paper
\makeatletter
\AfterEndEnvironment{algorithm}{\let\@algcomment\relax}
\AtEndEnvironment{algorithm}{\kern2pt\hrule\relax\vskip3pt\@algcomment}
\let\@algcomment\relax
\newcommand\algcomment[1]{\def\@algcomment{\footnotesize#1}}
\renewcommand\fs@ruled{\def\@fs@cfont{\bfseries}\let\@fs@capt\floatc@ruled
  \def\@fs@pre{\hrule height.8pt depth0pt \kern2pt}%
  \def\@fs@post{}%
  \def\@fs@mid{\kern2pt\hrule\kern2pt}%
  \let\@fs@iftopcapt\iftrue}
\makeatother

\newcommand{\xmark}{\ding{55}}

\newcolumntype{x}[1]{>{\centering\arraybackslash}p{#1pt}}
\newcolumntype{y}[1]{>{\raggedright\arraybackslash}p{#1pt}}
\newcolumntype{z}[1]{>{\raggedleft\arraybackslash}p{#1pt}}
\newlength\savewidth\newcommand\shline{\noalign{\global\savewidth\arrayrulewidth  \global\arrayrulewidth 1pt}\hline\noalign{\global\arrayrulewidth\savewidth}}
\newcommand{\tablestyle}[2]{\setlength{\tabcolsep}{#1}\renewcommand{\arraystretch}{#2}\centering\footnotesize}

\definecolor{Highlight}{HTML}{39b54a}  % green
\definecolor{green}{HTML}{39b54a} % more transparent
\definecolor{red}{HTML}{cb4335} % red

\DeclareCaptionLabelFormat{andtable}{#1~#2  \&  \tablename~\thetable}

\title{Revisiting Contrastive Methods for Unsupervised Learning of Visual Representations}

\author{
\hspace{-1.0em} Wouter Van Gansbeke\thanks{Equal contribution} $^{ , 1}$ \quad \hspace{-0.6em} Simon Vandenhende\footnotemark[1] $^{ , 1}$ \quad  \hspace{-0.6em} Stamatios Georgoulis$^2$ \quad  \hspace{-0.6em} Luc Van Gool$^{1,2}$  \\
\\
\vspace{-.2em}
\normalfont $^1$ KU Leuven/ESAT-PSI  \quad \normalfont $^2$ ETH Zurich/CVL}

\begin{document}

\maketitle

\begin{abstract}
Contrastive self-supervised learning has outperformed supervised pretraining on many downstream tasks like segmentation and object detection. However, current methods are still primarily applied to curated datasets like ImageNet. In this paper, we first study how biases in the dataset affect existing methods. Our results show that an approach like MoCo~\cite{he2019momentum} works surprisingly well across: (i) object- versus scene-centric, (ii) uniform versus long-tailed and (iii) general versus domain-specific datasets. Second, given the generality of the approach, we try to realize further gains with minor modifications. We show that learning additional invariances - through the use of multi-scale cropping, stronger augmentations and nearest neighbors - improves the representations. Finally, we observe that MoCo learns spatially structured representations when trained with a multi-crop strategy. The representations can be used for semantic segment retrieval and video instance segmentation without finetuning. Moreover, the results are on par with specialized models. We hope this work will serve as a useful study for other researchers. The code and models are available~\footnote{Code: \url{https://github.com/wvangansbeke/Revisiting-Contrastive-SSL}}.
\end{abstract}

\section{Introduction}
\label{sec: introduction}
Self-supervised learning (SSL)~\cite{jing2020self} aims to learn powerful representations without relying on human annotations. The representations can be used for various purposes, including transfer learning~\cite{he2019momentum}, clustering~\cite{asano2020labelling,van2020scan,van2021unsupervised} or semi-supervised learning~\cite{chen2020big}. Recent self-supervised methods~\cite{chen2020simple,he2019momentum,caron2020unsupervised,grill2020bootstrap} learn visual representations by imposing invariances to various data transformations. A popular way of formulating this idea is through the instance discrimination task~\cite{wu2018unsupervised} - which treats each image as a separate class. Augmentations of the same image are considered as positive examples of the class, while other images serve as negatives. To handle the large number of instance classes, the task is expressed as a non-parametric classification problem using the contrastive loss~\cite{gutmann2010noise,oord2018representation}. 

Despite the recent progress, most methods still train on images from ImageNet~\cite{deng2009imagenet}. This dataset has specific properties: (1) the images depict a single object in the center of the image, (2) the classes follow a uniform distribution and (3) the images have discriminative visual features. To deploy self-supervised learning into the wild, we need to quantify the dependence on these properties. Therefore, in this paper, we first study the influence of dataset \emph{biases} on the representations. We take a utilitarian view and transfer different representations to a variety of downstream tasks.

Our results indicate that an approach like MoCo~\cite{he2019momentum} works well for both object- and scene-centric datasets. We delve deeper to understand these results. A key component is the augmentation strategy which involves random cropping. For an object-centric dataset like ImageNet, two crops from the same image will show a portion of the same object and no other objects. However, when multiple objects are present, a positive pair of non-overlapping crops could lead us to wrongfully match the feature representations of different objects. This line of thought led recent studies~\cite{purushwalkam2020demystifying,selvaraju2020casting} to believe that contrastive SSL benefits from object-centric data. 

So how can contrastive methods learn useful representations when applied to more complex, scene-centric images? We propose a hypothesis that is two-fold. First, the default parameterization of the augmentation strategy avoids non-overlapping views. As a result, positive pairs will share information, which means that we can match their representations. Second, when applying more aggressive cropping, we only observe a small drop in the transfer learning performance. Since patches within the same image are strongly correlated, maximizing the agreement between non-overlapping views still provides a useful learning signal. We conclude that, in visual pretraining, combining the instance discrimination task with random cropping is universally applicable. 

The common theme of recent advances is to learn representations that are invariant to different transformations. Starting from this principle, we try to improve the results obtained with an existing framework~\cite{he2019momentum}. More specifically, we investigate three ways of generating a more diverse set of positive pairs. First, the multi-crop transform from~\cite{caron2020unsupervised} is revisited. Second, we examine the use of a stronger augmentation policy. Third, we leverage nearest neighbors mined online during training as positive views. The latter imposes invariances which are difficult to learn using handcrafted image transformations. The proposed implementation requires only a few lines of code and provides a simple, yet effective alternative to clustering based methods~\cite{caron2020unsupervised,li2020prototypical}. Each of the proposed additions is found to boost the performance of the representations under the transfer learning setup. 

The multi-crop transform realizes significant gains. We probe into what the network learns to explain the improvements. The multi-crop transform maximizes the agreement between smaller (local) crops and a larger (global) view of the image. This forces the model to learn a more spatially structured representation of the scene. As a result, the representations can be directly used to solve several dense prediction tasks without any finetuning. In particular, we observe that the representations already model class semantics and dense correspondences. Furthermore, the representations are competitive with specialized methods~\cite{jabri2020walk,zhang2020self}. In conclusion, the multi-crop setup provides a viable alternative to learn dense representations without relying on video data~\cite{jabri2020walk,lai2020mast} or handcrafted priors~\cite{hwang2019segsort,zhang2020self,van2021unsupervised}.

In summary, the overall goal of this paper is to learn more effective representations through contrastive self-supervised learning without relying too much on specific dataset biases. The remainder of this paper is structured as follows. Section~\ref{sec: framework} introduces the framework.
Section~\ref{sec: contrastive_wild} whether the standard SimCLR augmentations transfer across different datasets. This question is answered positively. Based upon this result, Section~\ref{sec: invariances} then studies the use of additional invariances to further improve the learned representations. We hope this paper will provide useful insights to other researchers. .
\section{Framework}
\label{sec: framework}
We briefly introduce the contrastive learning framework. The idea is to generate feature representations that maximize the agreement between similar (\emph{positive}) images and minimize the agreement between dissimilar (\emph{negative}) images. Let $x$ be an image. Assume that a set of positives for $x$ can be acquired, denoted by $\mathcal{X}^+$. Similarly, a set of negatives $\mathcal{X}^-$ is defined. We learn an embedding function $f$ that maps each sample on a normalized hypersphere. The contrastive loss~\cite{gutmann2010noise,oord2018representation} takes the following form
\begin{equation}
\mathcal{L}_{\text{contrastive}} = -\sum_{x^+ \in \mathcal{X^+}} \log \frac{\exp\left[f(x)^T \cdot f(x^+) / \tau\right]}{\exp\left[ f(x)^T \cdot f(x^+) / \tau\right] + \sum_{x^- \in\mathcal{X^-}} \exp\left[f(x)^T \cdot f(x^-) / \tau\right]}
\label{eq: contrastive_loss}
\end{equation}
where $\tau$ is a temperature hyperparameter. We will further refer to the image $x$ as the \emph{anchor}. 

SSL methods obtain positives and negatives by treating each image as a separate class~\cite{wu2018unsupervised}. More specifically, augmented views of the same image are considered as positives, while other images are used as negatives. The data augmentation strategy is an important design choice as it determines the invariances that will be learned. Today, most works rely on a similar set of augmentations that consists of (1) cropping, (2) color distortions, (3) horizontal flips and (4) Gaussian blur.

In this paper, we build upon MoCo~\cite{he2019momentum} - a widely known and competitive framework. However, our findings are expected to apply to other related methods (e.g. SimCLR~\cite{chen2020simple}) as well. The embedding function $f$ with parameters $\theta_f$ consists of a backbone $g$ (e.g. ResNet~\cite{he2016deep}) and a projection MLP head $h$. The contrastive loss is applied after the projection head $h$. MoCo uses a queue and a moving-averaged encoder $f'$ to keep a large and consistent set of negative samples. The parameters $\theta_{f'}$ of $f'$ are updated as: $\theta_{f'} = m \theta_{f'} + (1-m) \theta_f$ with $m$ a momentum hyperparameter. The momentum-averaged encoder $f'$ takes as input the anchor $x$, while the encoder $f$ is responsible for the positives $\mathcal{X}^+$. The queue maintains the encoded anchors as negatives. We refer to~\cite{he2019momentum} for more details. 
\section{Contrastive Learning in the Wild}
\label{sec: contrastive_wild}
Most contrastive self-supervised methods train on unlabeled images from ImageNet~\cite{deng2009imagenet}. This is a curated dataset with unique characteristics. First, the images are \emph{object-centric}, i.e. they depict only a single object in the center of the image. This differs from other datasets~\cite{lin2014microsoft,kuznetsova2020open} which contain more complex scenes with several objects. Second, the underlying classes are \emph{uniformly} distributed. Third, the images have \emph{discriminative} visual features. For example, ImageNet covers various bird species which can be distinguished by a few key features. In contrast, domain-specific datasets (e.g. BDD100K~\cite{yu2020bdd100k}) contain less discriminative scenery showing the same objects like cars, pedestrians, etc. In this section, we study the influence of dataset biases for contrastive self-supervised methods. 

\paragraph{Setup.} We train MoCo-v2~\cite{chen2020improved} on a variety of datasets. Table~\ref{tab: datasets} shows an overview. The representations are evaluated on six downstream tasks: linear classification, semantic segmentation, object detection, video instance segmentation and depth estimation. We adopt the following target datasets for linear classification: CIFAR10~\cite{krizhevsky2009learning}, Food-101~\cite{krause2013collecting}, Pets~\cite{parkhi2012cats}, Places365~\cite{zhou2017places}, Stanford Cars~\cite{krause2013collecting}, SUN397~\cite{xiao2010sun} and VOC 2007~\cite{everingham2010pascal}. The semantic segmentation task is evaluated on Cityscapes~\cite{cordts2016cityscapes}, PASCAL VOC~\cite{everingham2010pascal} and NYUD~\cite{silberman2012indoor}. We use PASCAL VOC~\cite{everingham2010pascal} for object detection. The DAVIS-2017 benchmark~\cite{davis2017} is used for video instance segmentation. Finally, depth estimation is performed on NYUD~\cite{silberman2012indoor}. The model, i.e. a ResNet-50 backbone, is pretrained for 400 epochs using batches of size 256. The initial learning rate is set to $0.3$ and decayed using a cosine schedule. We use the default values for the temperature ($\tau=0.2$) and momentum ($m=0.999$) hyperparameters. 

%%%%%% Datasets overview
\newcommand{\tabtop}[1]{\fontsize{7.5pt}{1em}\selectfont \textbf{#1}}
\newcommand{\tabelem}[1]{{#1}}
\begin{table}
\caption{\textbf{Overview of the training datasets.} We sample a uniform and long-tailed (LT) subset of 118K images from ImageNet. On OpenImages, we sample a random subset of 118K images. The complete train splits are used for COCO and BDD100K. The figure shows some examples.}
\label{tab: datasets}
\begin{minipage}{0.65\textwidth}
\centering
\tablestyle{2.0pt}{1.1}
    \begin{tabular}{lccccc}
    \toprule
    \tabtop{Pretrain Data} & ~ & \tabtop{\#Imgs} & \tabtop{\#Obj/Img} & \tabtop{Uniform} & \tabtop{Discriminative} \\
    \hline 
    \tabelem{ImageNet-118K~\cite{deng2009imagenet}} & ~ & \tabelem{118 K} & \tabelem{1.7} & \tabelem{$\checkmark$} & $\checkmark$ \\
    \tabelem{ImageNet-118K-LT~\cite{deng2009imagenet}} & ~ & \tabelem{118 K} & \tabelem{1.7} & \tabtop{\xmark} & \tabelem{$\checkmark$} \\
    \tabelem{COCO~\cite{lin2014microsoft}} & ~ & \tabelem{118 K} & \tabelem{7.3} & \tabtop{\xmark} & \tabelem{$\checkmark$} \\
    \tabelem{OpenImages-118K~\cite{kuznetsova2020open}} & ~ & \tabelem{118 K} & \tabelem{8.4} & \tabtop{\xmark} & \tabelem{$\checkmark$} \\
    \tabelem{BDD100K~\cite{yu2020bdd100k}} & ~ & \tabelem{90 K} & - & \tabtop{\xmark} & \tabtop{\xmark} \\
    \bottomrule
    \end{tabular}
\end{minipage}
\begin{minipage}{0.32\textwidth}
\centering
\vspace{-.85em}
\includegraphics[width=1.0\linewidth]{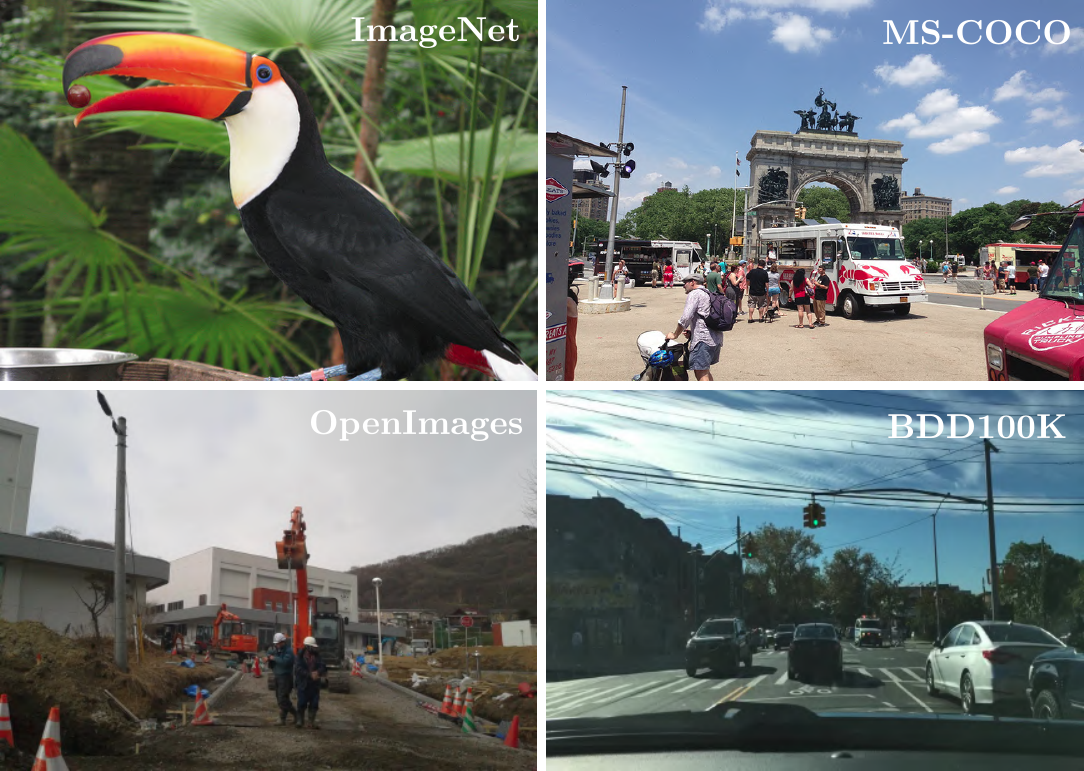}
\end{minipage}
\end{table}

%%%%% Table commands
\newcommand{\hr}[1]{\textcolor{red}{#1}}
\newcommand{\hg}[1]{\textcolor{green}{#1}}

\newcommand{\ressup}[2]{\tablestyle{1pt}{1} 
\begin{tabular}{z{16}y{18}} {#1} & {} \end{tabular}}

\newcommand{\ressupnyud}[2]{\tablestyle{1pt}{1} 
\begin{tabular}{z{20}y{28}} {#1} & {} \end{tabular}}

\newcommand{\res}[3]{
\tablestyle{1pt}{1}
\begin{tabular}{z{16}y{18}}
{#1} & \fontsize{7.0pt}{1em}\selectfont{~(${#2}${#3})}
\end{tabular}}

\newcommand{\reshg}[3]{
\tablestyle{1pt}{1} 
\begin{tabular}{z{16}y{18}}
{#1} &
\fontsize{7.0pt}{1em}\selectfont{~\hg{(${#2}$\textbf{#3})}}
\end{tabular}}

\newcommand{\reshgnyud}[3]{
\tablestyle{1pt}{1} 
\begin{tabular}{z{20}y{28}}
{#1} &
\fontsize{7.0pt}{1em}\selectfont{~\hg{(${#2}$\textbf{#3})}}
\end{tabular}}

\newcommand{\reshn}[3]{
\tablestyle{1pt}{1} 
\begin{tabular}{z{16}y{18}}
{#1} &
\fontsize{7.0pt}{1em}\selectfont{~(${#2}$\textbf{#3})}
\end{tabular}}

\newcommand{\reshr}[3]{
\tablestyle{1pt}{1} 
\begin{tabular}{z{16}y{18}}
{#1} &
\fontsize{7.0pt}{1em}\selectfont{~\hr{(${#2}$\textbf{#3})}}
\end{tabular}}

\newcommand{\demph}[1]{\textcolor{Gray}{#1}}
\newcommand{\resrand}[2]{\tablestyle{1pt}{1} \begin{tabular}{z{16}y{18}} \demph{#1} & {} \end{tabular}}

%%%%%% Evaluation
\begin{table}[!t]
    \vspace{-1.5em}
    \label{tab: contrastive_in_the_wild}
    \tablestyle{.8pt}{1.1}
    \centering
    \caption{Comparison of linear classification models trained on top of frozen features (400 epochs pretraining).}
    %\vspace{0.5em}
    \label{tab: linear_classification}
    \begin{tabular}{y{46}|x{46}|x{46}|x{46}|x{46}|x{46}|x{46}|x{46} c}
    \toprule
    \tabtop{Pretrain Data} & \tabtop{CIFAR10} & \tabtop{Cars} & \tabtop{Food} & \tabtop{Pets} & \tabtop{Places} & \tabtop{SUN} & \tabtop{VOC} \\
    \hline
    \tabelem{IN-118K} & \ressup{83.1}{} & \ressup{35.9}{} & \ressup{62.2}{} & \ressup{68.9}{} & \ressup{45.0}{} & \ressup{50.0}{} & \ressup{75.8}{} & ~\\
    \tabelem{COCO} & \reshr{77.4}{-}{5.7} & \reshr{33.9}{-}{2.0} & \reshr{62.0}{-}{0.2} & \reshr{62.6}{-}{6.3} & \reshg{47.3}{+}{2.3} & \reshg{53.6}{+}{3.6} & \reshg{80.9}{+}{5.1} & ~\\
    \tabelem{OI-118K} & \reshr{74.0}{-}{9.1} & \reshr{32.2}{-}{3.7} & \reshr{58.4}{-}{3.8} & \reshr{59.3}{-}{9.6} & \reshg{46.6}{+}{1.6} & \reshg{52.3}{+}{2.3} & \reshg{75.9}{+}{0.1} & ~\\
    \tabelem{IN-118K-LT} & \reshg{83.2}{+}{0.1} & \reshg{36.1}{+}{0.2} & \reshr{62.1}{-}{0.1} & \reshg{69.1}{+}{0.2} & \reshg{45.3}{+}{0.3} & \reshg{50.4}{+}{0.4} & \reshg{76.1}{+}{0.3} &  ~\\
    \bottomrule
    \end{tabular}
\end{table}

\begin{table}[!t]
    \vspace{-1.5em}
    \caption{Comparison of different representations under the transfer learning setup (400 epochs pretraining).}
    %\vspace{0.5em}
    \label{tab: transfer_learning}
    \tablestyle{.8pt}{1.1}
    \centering
    \begin{tabular}{ry{28}|x{44}|x{44}|x{44}|x{50}|x{62}|x{50} c}
    \toprule
    ~ & ~ & \multicolumn{3}{c|}{\fontsize{7.5pt}{1em}\selectfont \textbf{Semantic seg. } (mIoU)} &
    {\fontsize{7.5pt}{1em}\selectfont \textbf{Detection} (AP)} &
    {\fontsize{7.5pt}{1em}\selectfont \textbf{Vid. seg.~ ($\mathcal{J}\&\mathcal{F})$}} & {\fontsize{7.5pt}{1em}\selectfont \textbf{Depth} (rmse)} & \\
    \multicolumn{2}{l|}{{\fontsize{7.5pt}{1em}\selectfont \textbf{Pretrain Data}}} &
    {\fontsize{7.5pt}{1em}\selectfont \textbf{VOC}} &
    {\fontsize{7.5pt}{1em}\selectfont \textbf{Cityscapes}} & 
    {\fontsize{7.5pt}{1em}\selectfont \textbf{NYUD}} &
    {\fontsize{7.5pt}{1em}\selectfont \textbf{VOC}} &
    {\fontsize{7.5pt}{1em}\selectfont \textbf{DAVIS}} &  
    {\fontsize{7.5pt}{1em}\selectfont \textbf{NYUD}} & \\
    \hline
    \multicolumn{2}{l|}{IN-118K} & \ressup{68.9}{} & \ressup{70.1}{} & \ressup{37.7}{} & \ressup{53.0}{} & \ressup{63.5}{} & \ressupnyud{0.625}{} & \\
    \multicolumn{2}{l|}{COCO} & \reshg{69.1}{+}{0.2} & \reshg{70.3}{+}{0.2} & \reshg{39.3}{+}{1.6} & \res{53.0}{+}{0.0} & \reshg{65.1}{+}{1.6} & \reshgnyud{0.612}{-}{0.013} & \\
    \multicolumn{2}{l|}{OI-118K} & \reshr{67.9}{-}{1.0} & \reshg{70.9}{+}{0.8} & \reshg{38.4}{+}{0.7} & \reshg{53.1}{+}{0.1} & \reshg{64.8}{+}{1.3} & \reshgnyud{0.609}{-}{0.016} & \\
    \multicolumn{2}{l|}{IN-118K-LT} & \reshg{69.0}{+}{0.1} & \res{70.1}{+}{0.0} & \reshg{37.9}{+}{0.2} & \res{53.0}{+}{0.0} & \reshg{63.7}{+}{0.2} & \reshgnyud{0.622}{-}{0.003} & \\
    \multicolumn{2}{l|}{BDD100K} & - & \res{70.1}{+}{0.0} & - & - & - & - & ~\\
    \bottomrule
    \end{tabular}
\end{table}

\subsection{Object-centric Versus Scene-centric}
\label{subsec: subsec_contrastive_wild_object_scene}
We compare the representations trained on an object-centric dataset - i.e. ImageNet (IN-118K) - against the ones obtained from two scene-centric datasets - i.e. COCO and OpenImages (OI-118K). Tables~\ref{tab: linear_classification}-\ref{tab: transfer_learning} show the results under the linear classification and transfer learning setup. 

\paragraph{Results.} The linear classification model yields better results on CIFAR10, Cars, Food and Pets when pretraining the representations on ImageNet. Differently, when tackling the classification task on Places, SUN and VOC, the representations from COCO and OpenImages are better suited. The first group of target benchmarks contains images centered around a single object, while the second group contains scene-centric images with multiple objects. We conclude that, for linear classification, the pretraining dataset should match the target dataset in terms of being object- or scene-centric.

Next, we consider finetuning. Perhaps surprisingly, we do not observe any significant disadvantages when using more complex images from COCO or OpenImages for pretraining. In particular, for the reported tasks, the COCO pretrained model even outperforms its ImageNet counterpart. We made a similar observation when increasing the size of the pretraining dataset (see suppl. materials). 

\begin{figure}[t]
\begin{minipage}[b]{.32\linewidth} % FIG 1
\centering
\includegraphics[width=\textwidth]{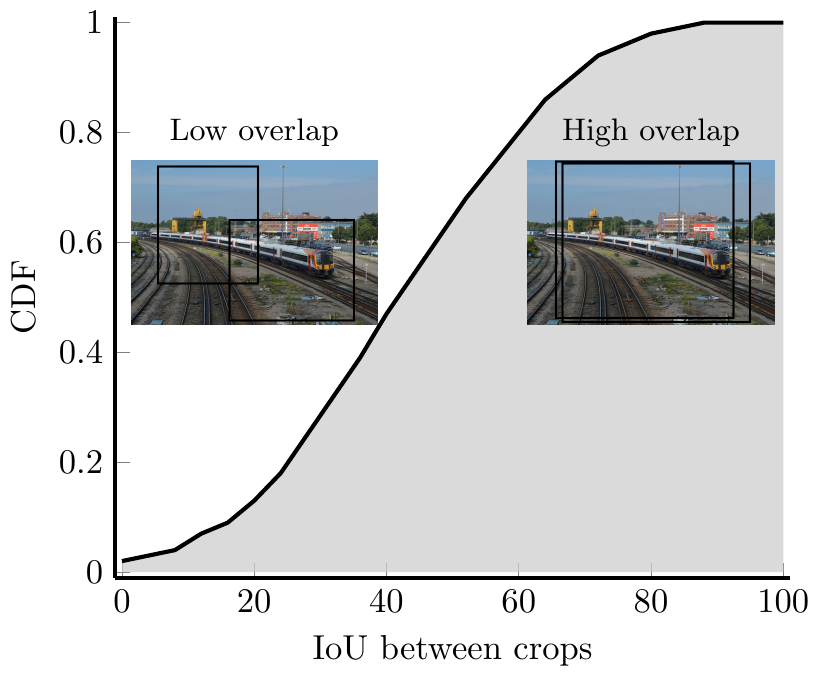}
\caption{IoU between random resized crops for existing frameworks.}
\label{fig: area_histogram}
\end{minipage} %
\hspace{0.01\linewidth}
\begin{minipage}[b]{.32\linewidth} % FIG 2
\centering
\includegraphics[width=\textwidth]{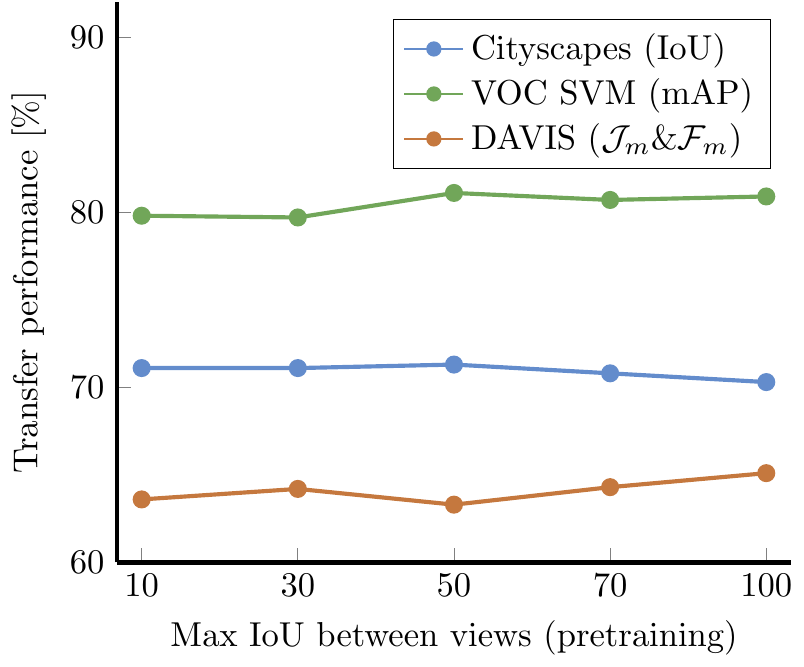}
\caption{Transfer results when thresholding the IoU between crops.}
\label{fig: transfer_iou}
\end{minipage} %
\hspace{0.01\linewidth}
\begin{minipage}[b]{.32\linewidth} % FIG 3
\centering
\includegraphics[width=\textwidth]{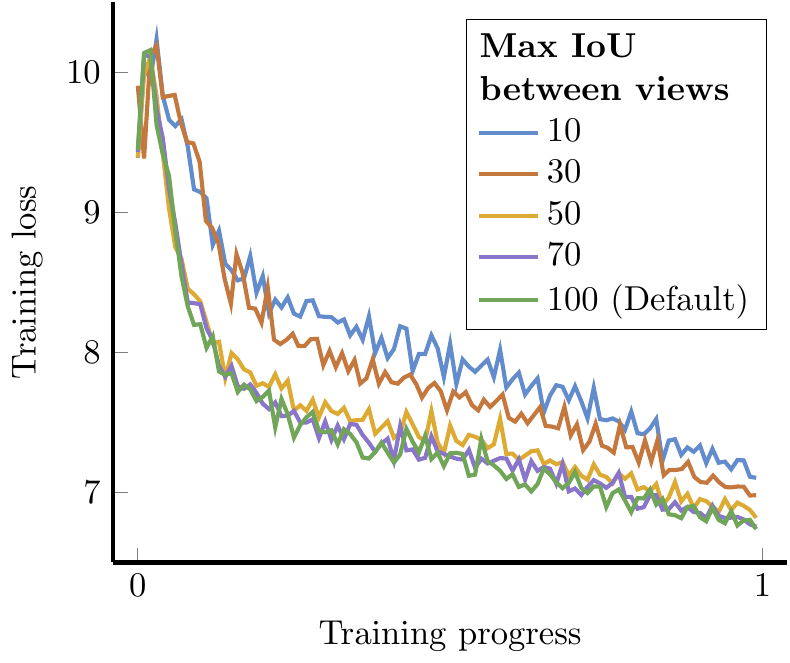}
\caption{Training curves when thresholding the IoU between crops.}
\label{fig: loss_iou}
\end{minipage} %
\end{figure}

\paragraph{Discussion.} In contrast to prior belief~\cite{purushwalkam2020demystifying,selvaraju2020casting}, our results indicate that contrastive self-supervised methods do not suffer from pretraining on scene-centric datasets. How can we explain this inconsistency with earlier studies? First, the  experimental setup in~\cite{purushwalkam2020demystifying} only considered the linear evaluation protocol for an object-centric dataset (i.e. PASCAL cropped boxes). This analysis~\cite{purushwalkam2020demystifying} does not show us the full picture. Second, the authors conclude that existing methods suffer from using scene-centric datasets due to the augmentation strategy, which involves random cropping. They argue that aggressive cropping could yield non-overlapping views which contain different objects. In this case, maximizing the feature similarity would be detrimental for object recognition tasks. However, the default cropping strategy barely yields non-overlapping views\footnote{We use the \texttt{RandomResizedCrop} in PyTorch with scaling $s=(0.2,1.0)$ and output size $224\times224$.}. This is verified by plotting the intersection over union (IoU) between pairs of crops (see Figure~\ref{fig: area_histogram}). 
We conclude that the used example of non-overlapping crops~\cite{purushwalkam2020demystifying,selvaraju2020casting} seldom occurs. 

The above observation motivates us to reconsider the importance of using overlapping views. We pretrain on COCO while forcing the IoU between views to be smaller than a predefined threshold. Figures~\ref{fig: transfer_iou}-\ref{fig: loss_iou} show the transfer performance and training curves for different values of the threshold. The optimization objective is harder to satisfy when applying more aggressive cropping (i.e. the training loss increases when lowering the IoU). However, Figure~\ref{fig: transfer_iou} shows that the transfer performance remains stable. Patches within the same image were observed at the same point in time and space, which means that they will share information like the camera viewpoint, color, shape, etc. As a result, the learning signal is still meaningful, even when less overlapping crops are used as positives.

\subsection{Uniform Versus Long-tailed}
\label{subsec: subsec_contrastive_wild_uniform}
Next, we study whether MoCo benefits from using a uniform (IN-118K) versus long-tailed (IN-118K-LT) dataset. We adopt the sampling strategy from~\cite{liu2019large} to construct a long-tailed version of ImageNet. The classes follow the Pareto distribution with power value $\alpha=6$. Tables~\ref{tab: linear_classification}-\ref{tab: transfer_learning} indicate that MoCo is robust to changes in the class distribution of the dataset. In particular, the IN-118K-LT model performs on par or better compared to its IN-118K counterpart across all tasks. We conclude that it is not essential to use a uniformly distributed dataset for pretraining. 

\subsection{Domain-Specific Datasets}
\label{subsec: domain_specific}
Images from ImageNet have discriminative visual features, e.g. the included bird species can be recognized from their beak or plumage. In this case, the model could achieve high performance on the instance discrimination task by solely focusing on the most discriminative feature in the image. Differently, in urban scene understanding, we deal with more monotonous scenery - i.e. all images contain parked cars, lane markings, pedestrians, etc. We try to measure the impact of using less discriminative images for representation learning. 

We pretrain on BDD100K - a dataset for urban scene understanding. Table~\ref{tab: transfer_learning} compares the representations for the semantic segmentation task on Cityscapes when pretraining on BDD100K versus IN-118K. The BDD100K model performs on par with the IN-118K pretrained baseline (BDD100K: $+0.00 \%$). This shows that MoCo can be applied to domain-specific data as well. Two models trained on very different types of data perform equally well. A recent study~\cite{zhao2020makes} showed that it is mostly the low- and mid-level visual features that are retained under the transfer learning setup. This effect explains how two different representations produce similar results. Finally, we expect that further gains can be achieved by finetuning the augmentation strategy using domain knowledge. 
\section{Learning Invariances}
\label{sec: invariances}
In Section~\ref{sec: contrastive_wild}, we showed that MoCo works well for a large variety of pretraining datasets: scene-centric, long-tailed and domain-specific images. This result shows that there is no obvious need to use a more advanced pretext task~\cite{selvaraju2020casting} or use an alternative source of data like video~\cite{purushwalkam2020demystifying} to improve the representations. Therefore, instead of developing a new framework, we try to realize further gains while sticking with the approach from MoCo. We realize this objective by learning additional invariances. We study three such mechanisms, i.e. multi-scale constrained cropping, stronger augmentations and the use of nearest neighbors. We concentrate on the implementation details and uncover several interesting qualities of the learned representations. Table~\ref{tab: overview_invariances} shows an overview of the results. 

%%%% Overview table
\definecolor{darkF7E0D5}{RGB}{0,0,0}
\newcommand{\rownumber}[1]{\textcolor{darkF7E0D5}{#1}}
\renewcommand{\demph}[1]{\textcolor{Gray}{#1}}
\newcommand{\std}[1]{{\fontsize{5pt}{1em}\selectfont ~~$_\pm$$_{\text{#1}}$}}

\renewcommand{\hl}[1]{\textcolor{Highlight}{#1}}

\renewcommand{\res}[3]{
\tablestyle{1pt}{1}
\begin{tabular}{z{16}y{18}}
{#1} &
\fontsize{7.0pt}{1em}\selectfont{~(${#2}${#3})}
\end{tabular}}

\newcommand{\reshl}[3]{
\tablestyle{1pt}{1} 
\begin{tabular}{z{16}y{18}}
{#1} &
\fontsize{7.0pt}{1em}\selectfont{~\hl{(${#2}$\textbf{#3})}}
\end{tabular}}

\renewcommand{\resrand}[2]{\tablestyle{1pt}{1} \begin{tabular}{z{16}y{18}} \demph{#1} & {} \end{tabular}}
\renewcommand{\ressup}[2]{\tablestyle{1pt}{1} 
\begin{tabular}{z{16}y{18}} {#1} & {} \end{tabular}}

\begin{table}
\tablestyle{2pt}{1.1}
\caption{\textbf{Ablation of different components.} Models are pretrained for 200 epochs on COCO using the settings from Section~\ref{sec: contrastive_wild}. We indicate the differences with MoCo. The best model is trained with additional invariances.\vspace{0.5em}}
\label{tab: overview_invariances}

\begin{tabular}{ry{25}|x{12}x{12}x{12}x{12}x{12}|x{40}|x{40}|x{40}|x{40}|x{40}|x{40} c}
\toprule
\multicolumn{2}{c|}{} &
\multicolumn{5}{c|}{\fontsize{7.5pt}{1em}\selectfont \textbf{Setup} } &
\multicolumn{3}{c|}{\fontsize{7.5pt}{1em}\selectfont \textbf{Semantic seg. } (mIoU)} & 
\multicolumn{3}{c}{\fontsize{7.5pt}{1em}\selectfont \textbf{Classification} (mAP / Acc. / Acc.)}  
\\
\multicolumn{2}{l|}{Method} & 
{\fontsize{7.5pt}{1em}\selectfont \textbf{MC}} &
{\fontsize{7.5pt}{1em}\selectfont \textbf{CC}} &
{\fontsize{7.5pt}{1em}\selectfont \textbf{m$\downarrow$}} &
{\fontsize{7.5pt}{1em}\selectfont \textbf{A$^+$}} &
{\fontsize{7.5pt}{1em}\selectfont \textbf{NN}} &
{\fontsize{7.5pt}{1em}\selectfont \textbf{VOC}} &
{\fontsize{7.5pt}{1em}\selectfont \textbf{Cityscapes}} & 
{\fontsize{7.5pt}{1em}\selectfont \textbf{NYUD}} &
{\fontsize{7.5pt}{1em}\selectfont \textbf{VOC}} &
{\fontsize{7.5pt}{1em}\selectfont \textbf{ImageNet}} &
{\fontsize{7.5pt}{1em}\selectfont \textbf{Places}} &

\\
\hline
\multicolumn{2}{l|}{\demph{Rand. init.}}
& \demph{-} & \demph{-} & \demph{-} & \demph{-} & \demph{-}
& \resrand{39.2}{} 
& \resrand{65.0}{}  % city 
& \resrand{24.4}{}  % nyu
& \resrand{-}{}  % VOC 
& \resrand{-}{}  % IN
& \resrand{-}{}  % places
& \\
\multicolumn{2}{l|}{MoCo}
& \xmark & \xmark & \xmark & \xmark & \xmark
& \ressup{66.2}{} 
& \ressup{70.3}{}  % city mask
& \ressup{38.2}{}  % nyu
& \ressup{76.1}{}  % VOC svm
& \ressup{49.3}{}  % IN
& \ressup{45.1}{}  % places
& \\

\multicolumn{2}{l|}{\textcolor{red}{Sec.~\ref{subsec: multi_crop}}}
& \checkmark & \xmark & \xmark & \xmark & \xmark 
& \res{69.9}{+}{3.7}  % VOC sem
& \res{70.9}{+}{0.6}  % city
& \res{39.4}{+}{1.2}  % nyu
& \res{81.3}{+}{5.2}  % VOC svm
& \res{53.4}{+}{4.1}  % IN
& \res{47.7}{+}{2.6}  % places
& \\
\multicolumn{2}{l|}{}
& \checkmark & \checkmark & \xmark & \xmark & \xmark
& \res{70.2}{+}{4.0} 
& \res{70.9}{+}{0.6}  % city 
& \res{39.5}{+}{1.3}  % nyu
& \res{82.1}{+}{6.0}  % VOC svm
& \res{54.0}{+}{4.7}  % IN
& \res{47.9}{+}{2.8}  % places
& \\
\multicolumn{2}{l|}{}
& \checkmark & \checkmark & \checkmark & \xmark & \xmark
& \res{70.9}{+}{4.7} 
& \res{71.3}{+}{1.0}  % city
& \res{39.9}{+}{1.7}  % nyu
& \res{82.8}{+}{6.7}  % VOC svm
& \res{54.8}{+}{5.5}  % IN
& \res{48.1}{+}{3.0}  % places
& \\
\multicolumn{2}{l|}{\textcolor{red}{Sec.~\ref{subsec: auto_augment}}}
& \checkmark & \checkmark & \checkmark & \checkmark & \xmark
& \res{71.4}{+}{5.2} 
& \res{72.0}{+}{1.7}  % city
& \res{40.0}{+}{1.8}  % nyu
& \res{83.7}{+}{7.6}  % VOC svm
& \res{55.5}{+}{6.2}  % IN
& \res{48.2}{+}{3.1}  % places
& \\
\multicolumn{2}{l|}{\textcolor{red}{Sec.~\ref{subsec: neighbors}}}
& \checkmark & \checkmark & \checkmark & \checkmark & \checkmark
& \reshl{71.9}{+}{5.7} 
& \reshl{72.2}{+}{1.9}  % city
& \reshl{40.9}{+}{2.7}  % nyu
& \reshl{85.1}{+}{9.0}  % VOC svm
& \reshl{55.9}{+}{6.6}  % IN
& \reshl{48.5}{+}{3.4}  % places
& \\
\bottomrule
\multicolumn{14}{c}{\fontsize{7.5pt}{1em}\selectfont MC: Multi-crop, CC: Constrained multi-crop, m$\downarrow$: Lower momentum, A$^+$: Stronger augmentations, NN: Nearest neighbors}\\
\end{tabular}
\end{table}
\begin{figure}[!t]
    \begin{minipage}[t]{.62\textwidth}
    \includegraphics[width=\linewidth]{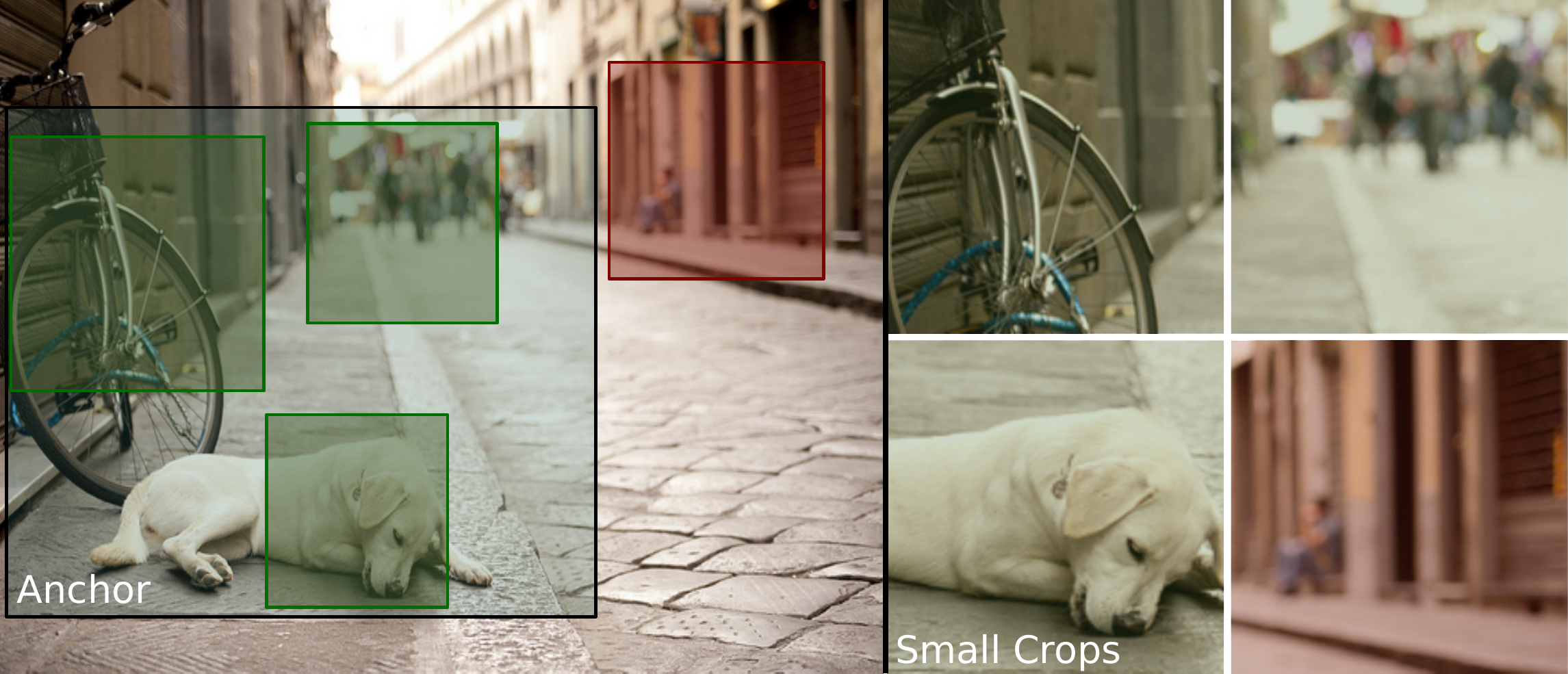}
    \captionof{figure}{The constrained multi-crop. Smaller crops are forced to lie within the anchor image. Invalid/valid crops are colored in red/green.}
    \label{fig: multiple_crops}
    \end{minipage} 
    \hfill
    \begin{minipage}[t]{.36\textwidth}
    \centering
    \includegraphics[width=0.85\linewidth]{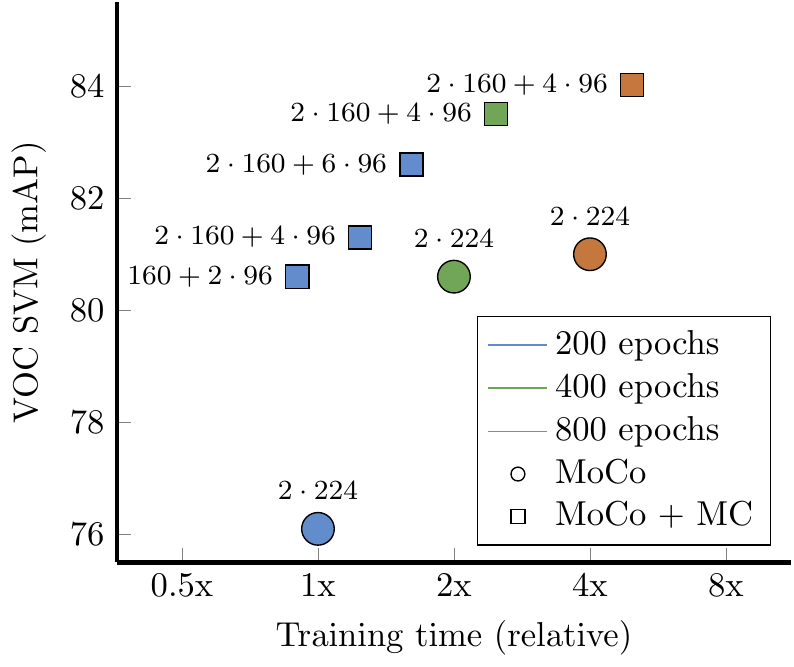}
    \caption{Transfer performance versus time for the multi-crop (MC) model.}
    \label{fig: multicrop_resource}
    \end{minipage}
\end{figure}

\subsection{Multi-Scale Constrained Cropping}
\label{subsec: multi_crop}
The employed cropping strategy proves crucial to obtain powerful visual representations. So far, we used the default two-crop transform, which samples two image patches to produce an anchor and a positive. The analysis in Section~\ref{sec: contrastive_wild} showed that this transformation yields highly-overlapping views. As a result, the model can match the anchor with its positive by attending only to the most discriminative image component, while ignoring other possibly relevant regions. This observation leads us to revisit the \texttt{multi-crop} augmentation strategy from~\cite{caron2020unsupervised}.

\paragraph{Setup.} The \texttt{multi-crop} transform samples $N$ additional positive views for each anchor. Remember, the views are generated by randomly applying resized cropping, color distortions, horizontal flipping and Gaussian blur. We adjust the scaling parameters and resolution of the random crop transform to limit the computational overhead. The two default crops (i.e. the anchor and a positive) contain between $20\%$ and $100\%$ of the original image area, and are resized to $160\times160$ pixels. Differently, the $N$ additional views contain between $5\%$ and $14\%$ of the original image area, and are resized to $96\times96$ pixels. We maximize the agreement between all $N+1$ positives and the anchor. Note that the $N$ smaller, more aggressively scaled crops are not used as anchors, i.e. they are not shown to the momentum-updated encoder and are not used as negatives in the memory bank. 

Further, we consider two additional modifications. First, we enforce the smaller crops to overlap with the anchor image. In this way, all positives can be matched with their anchor as there is shared information. The new transformation is referred to as \texttt{constrained multi-cropping}. Figure~\ref{fig: multiple_crops} illustrates the concept. Second, since increasing the number of views facilitates faster training, we reduce the momentum hyperparameter $m$ from $0.999$ to $0.995$. We pretrain for 200 epochs on COCO.

\paragraph{Results.} Table~\ref{tab: overview_invariances} benchmarks the  modifications. Both the \texttt{multi-crop} (MC) and \texttt{constrained multi-crop} (CC) significantly improve the transfer performance. Reducing the momentum (m$\downarrow$) yields further gains. Figure~\ref{fig: multicrop_resource} plots the training time versus performance for the \texttt{multi-crop} model. The \texttt{multi-crop} improves the computational efficiency w.r.t. the two-crop transform. 

\begin{table}[b]
\caption{\textbf{DAVIS 2017 video instance segmentation}. We use the publicly available code from~\cite{jabri2020walk} to evaluate our frozen representations. Qualitative results are shown for MoCo trained with the \texttt{multi-crop} (MC) transform.} 
\begin{minipage}{0.45\textwidth}
    \tablestyle{2pt}{1.1}
    \begin{tabular}{y{64} |x{44} | x{24} x{24} c}
    \toprule
    \fontsize{7.5pt}{1em}\selectfont \textbf{Method} &
    \fontsize{7.5pt}{1em}\selectfont \textbf{Data} &
    \fontsize{7.5pt}{1em}\selectfont $\mathcal{J}_m \uparrow$ & \fontsize{7.5pt}{1em}\selectfont $\mathcal{F}_m \uparrow$ & \\
    \hline
    \fontsize{7.5pt}{1em}\selectfont DenseCL~\cite{wang2020dense} &     \fontsize{7.5pt}{1em}\selectfont COCO &  60.6 & 63.9 & \\
    \fontsize{7.5pt}{1em}\selectfont MAST~\cite{lai2020mast} & \fontsize{7.5pt}{1em}\selectfont YT-VOS & 63.3 & 67.6 & \\
    \fontsize{7.5pt}{1em}\selectfont STC~\cite{jabri2020walk} & \fontsize{7.5pt}{1em}\selectfont Kinetics & 64.8 & 70.2 & \\
    \hline
    \fontsize{7.5pt}{1em}\selectfont MoCo & \fontsize{7.5pt}{1em}\selectfont COCO & 61.6 & 66.6 & \\
    \fontsize{7.5pt}{1em}\selectfont MoCo + MC & \fontsize{7.5pt}{1em}\selectfont COCO & 64.3 & 69.4 & \\
    \bottomrule
    \end{tabular}
\end{minipage}
\hfill
\begin{minipage}{0.55\textwidth}
\centering
\includegraphics[width=\linewidth]{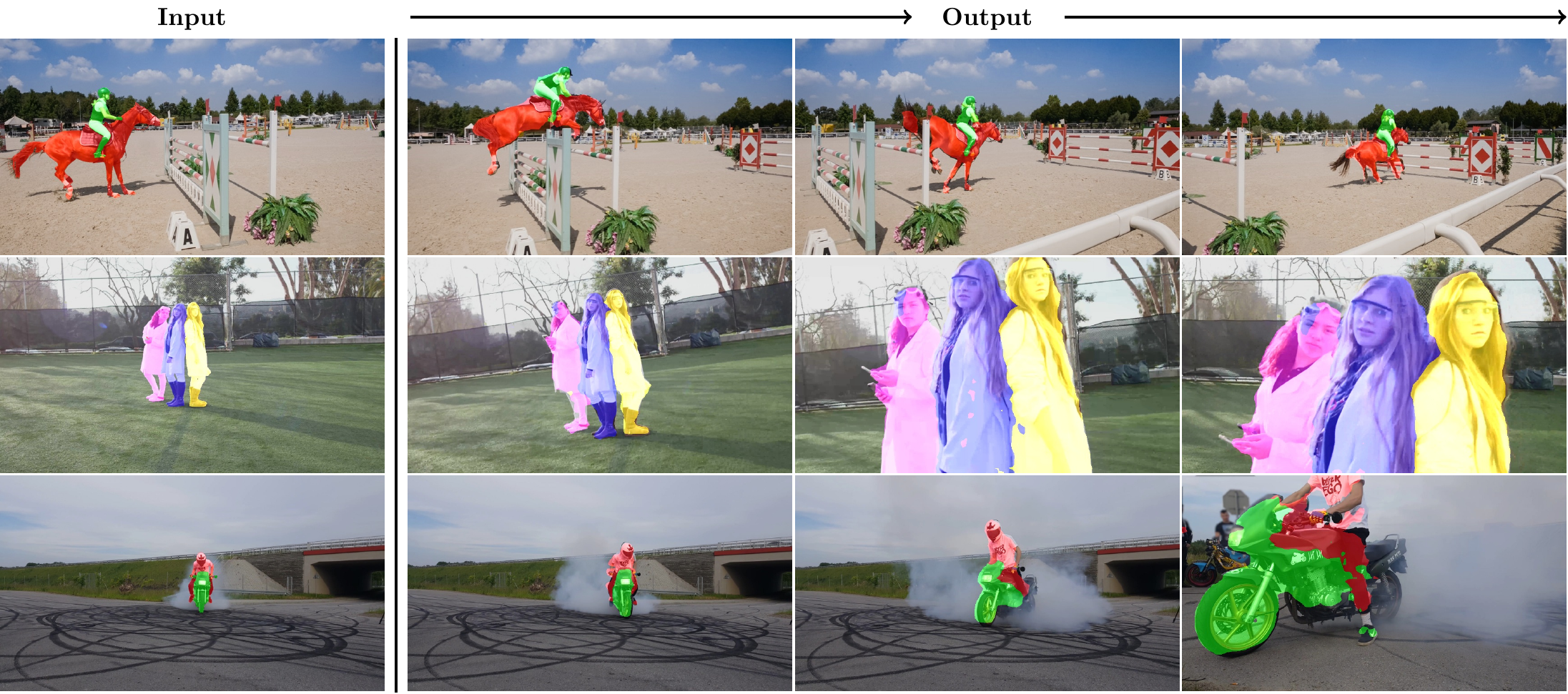}
\end{minipage}
\label{table: davis}
\end{table}

\paragraph{Discussion.} We investigate what renders the \texttt{multi-crop} and \texttt{constrained multi-crop} effective. The network matches the representations of the large anchor image with the small crops. Since the small crops show random subparts of the anchor image, the network is forced to encode as much information as possible, such that all image regions can be retrieved from the representation of the scene.
As a result, the representations will be informative of the spatial layout and different objects in the image - rather than attend only to the most discriminative image component.

\begin{wrapfigure}[14]{r}{0.32\textwidth}
%\vspace{-0.4em}
    \centering
    \includegraphics[width=1.0\linewidth]{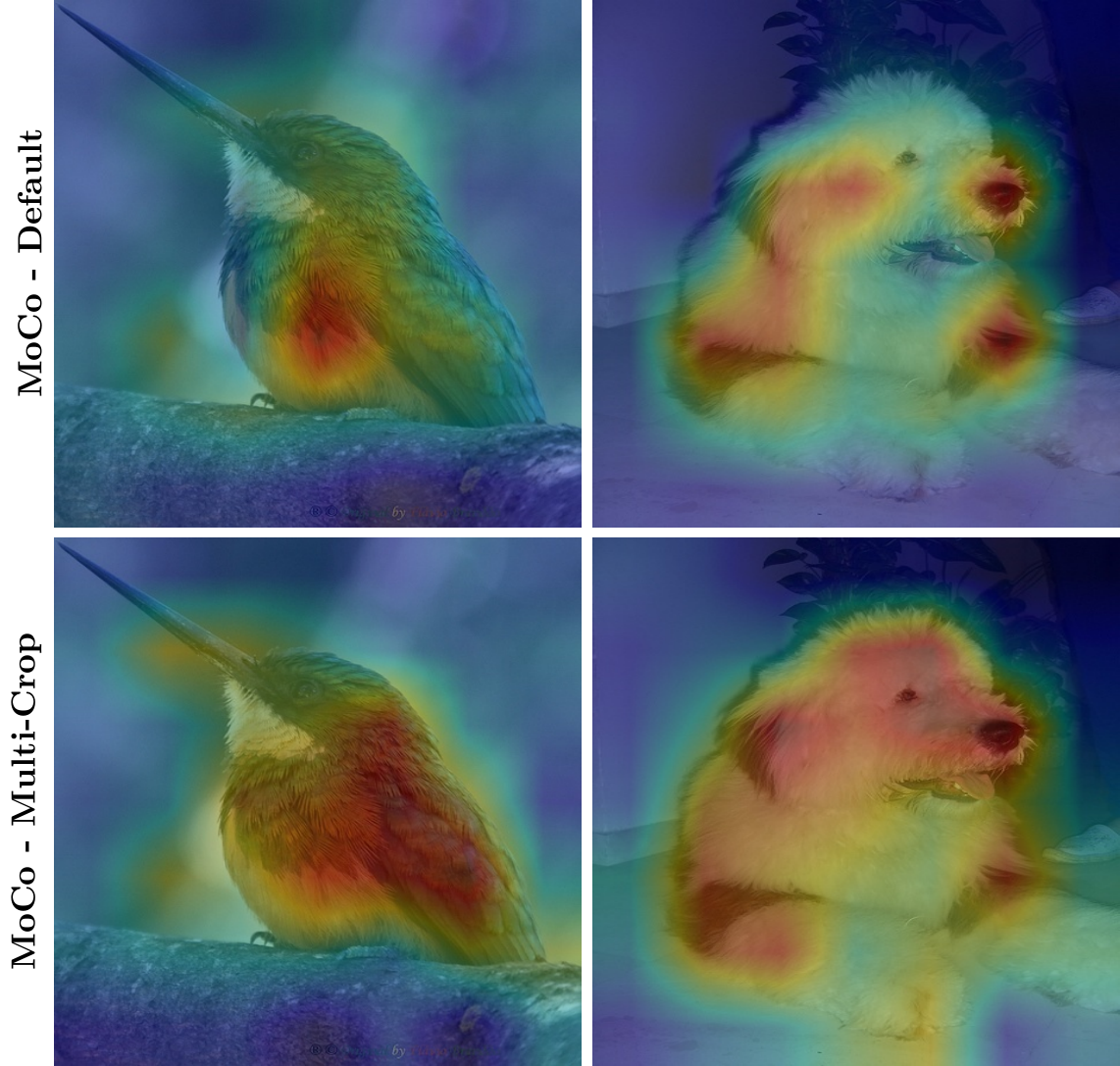}
    \captionof{figure}{Class activation maps for the two-crop vs. multi-crop model.}
    \label{fig: cams}
\end{wrapfigure}

The latter is verified by visualizing the class activation maps (CAMs)~\cite{zhou2016learning} of different representations (see Figure~\ref{fig: cams}). The CAMs of the \texttt{multi-crop} model segment the complete object, while the \texttt{two-crop} model only looks at a few discriminative components. Additional examples can be found in the suppl. materials.

Further, the representations are evaluated on two dense prediction tasks without any finetuning. We consider the tasks of video instance segmentation on DAVIS 2017~\cite{davis2017} and semantic segment retrieval on VOC~\cite{everingham2010pascal}. The \texttt{multi-crop} model is found superior at modeling dense correspondences for the video instance segmentation task in Table~\ref{table: davis} (MoCo vs. MoCo + MC). Similarly, Table~\ref{tab: voc} shows that the \texttt{multi-crop} model achieves higher performance on the semantic segment retrieval task compared to its two-crop counterpart. This indicates that the pixel embeddings are better disentangled according to the semantic classes. Finally, the MoCo \texttt{multi-crop} model is competitive with other methods that were specifically designed for the tasks. We conclude that the \texttt{multi-crop} setup provides a viable alternative to learn dense representations without supervision. Moreover, this setup does not rely on video~\cite{lai2020mast,jabri2020walk} or handcrafted priors~\cite{hwang2019segsort,zhang2020self,van2021unsupervised}.

\begin{table}[t]
\caption{\textbf{VOC semantic segment retrieval.} Each image is partitioned into a number of segments by running K-Means (K=15) on the spatial features. Then we adopt a region descriptor - computed by averaging the embeddings of all pixels within a segment - to obtain nearest neighbors of the validation regions from the train set. Qualitative results are shown for MoCo trained with the \texttt{multi-crop} (MC) augmentation strategy (K=5).}
\label{tab: voc}
\begin{minipage}{0.40\textwidth}
    \tablestyle{2pt}{1.1}
    \begin{tabular}{y{44}|x{52} x{54} c}
    \toprule
    & \multicolumn{2}{c}{\fontsize{7.5pt}{1em}\selectfont \textbf{Segm. Retr. VOC}} & \\
    \fontsize{7.5pt}{1em}\selectfont \textbf{Method} & \fontsize{7.5pt}{1em}\selectfont 7 classes (IoU)$\uparrow$ & \fontsize{7.5pt}{1em}\selectfont 21 classes (IoU)$\uparrow$ & \\
    \hline
    \fontsize{7.5pt}{1em}\selectfont SegSort~\cite{zhang2020self}
    & 10.2 & - & \\
    \fontsize{7.5pt}{1em}\selectfont SSL HG~\cite{zhang2020self}
    & 24.6 & - & \\
    \fontsize{7.5pt}{1em}\selectfont DenseCL~\cite{wang2020dense}
    & 48.4 & 35.1 & \\
    \hline
    \fontsize{7.5pt}{1em}\selectfont MoCo & 41.8 & 28.1 & \\
    \fontsize{7.5pt}{1em}\selectfont MoCo + MC & 48.1 & 35.1 & \\
    \bottomrule
    \end{tabular}
\end{minipage}
\hfill
\begin{minipage}{0.55\textwidth}
\centering
\includegraphics[width=1.0\linewidth]{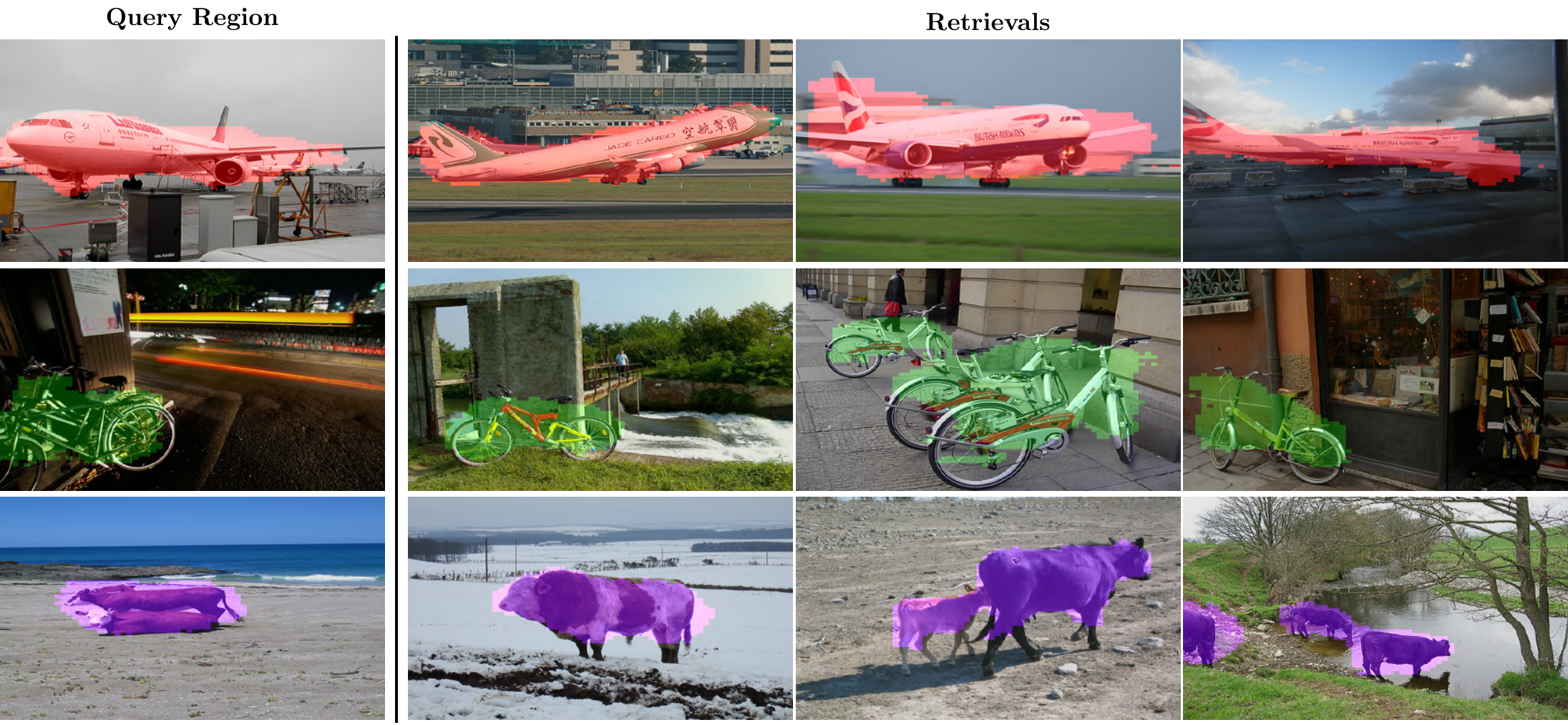}
\end{minipage}
\end{table}

\subsection{Stronger Augmentations}
\label{subsec: auto_augment}

As mentioned before, the image crops are augmented by random color distortions, horizontal flips and Gaussian blur. Can we do better by using stronger augmentations? We investigate the use of AutoAugment~\cite{cubuk2019autoaugment} - an advanced augmentation policy obtained using supervised learning on ImageNet. We consider three possible strategies to augment the positives: (1) standard augmentations, (2) AutoAugment and (3) randomly applying either (1) or (2). 

\begin{wrapfigure}[9]{r}{0.31\textwidth}
\begin{minipage}{\linewidth}
\vspace{-1em}
    \captionof{table}{Ablation of augmentation policies applied to random crops.}
    \tablestyle{2pt}{1.1}
    \centering
    \begin{tabular}{y{80}|x{35}}
    \toprule
    \fontsize{7.5pt}{1em}\selectfont \textbf{Augmentation policy} & 
    \fontsize{7.5pt}{1em}\selectfont \textbf{VOC SVM} (mAP)\\
    \hline
    \fontsize{7.5pt}{1em}\selectfont Standard & 82.8 \\
    \fontsize{7.5pt}{1em}\selectfont AutoAugment & 79.9 \\
    \fontsize{7.5pt}{1em}\selectfont Standard or AutoAugment & 83.7 \\
    \bottomrule
    \end{tabular}
    \label{tab: auto_augment_ablation}
\end{minipage}
\end{wrapfigure}
Table~\ref{tab: auto_augment_ablation} compares the representations under the linear evaluation protocol on PASCAL VOC. Replacing the standard augmentations with AutoAugment degrades the performance (from $82.8\%$ to $79.9\%$). However, randomly applying either the standard augmentations or AutoAugment does improve the result (from $82.8\%$ to $83.7\%$). Chen~\emph{et~al.}~\cite{chen2020simple} showed that contrastive SSL benefits from using strong color augmentations as two random crops from the same image will share a similar color distribution. AutoAugment applies fewer color distortions, resulting in lower performance. This is compensated when combining both augmentation strategies. Finally, Table~\ref{tab: overview_invariances} shows that combining our custom augmentation policy (A$^+$) with the model from Section~\ref{subsec: multi_crop} results in considerable improvements on all tasks. 

\subsection{Nearest Neighbors}
\label{subsec: neighbors}
Prior work~\cite{van2020scan} showed that the model learns to map visually similar images closer together than dissimilar ones when tackling the instance discrimination task. In this section, we build upon this insight by imposing additional invariances between neighboring samples. By leveraging other samples as positives, we can capture a rich set of deformations that are hard to model via handcrafted augmentations. However, a well-known problem with methods that group different samples together is preventing representation collapse. Therefore, we formulate our intuition as an auxiliary loss that regularizes the representations - keeping the instance discrimination task as the main force.

\paragraph{Setup.} Recall that the encoder $f$ consists of a backbone $g$ and a projection head $h$. MoCo maintains a queue $\mathcal{Q}_h$ of encoded anchors $\{q^{h}_0~...~q^{h}_{K-1}\}$ processed by the momentum encoder $f'$. We now introduce a second - equally sized and aligned - queue $\mathcal{Q}_g$ which maintains the features from before the projection head  $\{q^{g}_0~...~q^{g}_{K-1}\}$. The queue of backbone features $\mathcal{Q}_g$ can be used to mine nearest neighbors on-the-fly during training. In particular, for a positive $x^+$ - processed by the encoder $f$ -  the set of its $k$ nearest neighbors~$\mathcal{N}_{x^+} = \{q^h_i~|~sim(g(x^+),~q_i^g)~\text{is top}~k\in\mathcal{Q}_g\}$ is computed w.r.t. the queue $\mathcal{Q}_g$. The cosine similarity measure is denoted by $sim$. Finally, we use the contrastive loss to maximize the agreement between $x^+$ and its nearest neighbors $\mathcal{N}_{x^+}$ after the projection head:
\begin{equation}
    \label{eq: nn_loss}
    \mathcal{L}_{\text{nn}} = -\sum_{x^+ \in \mathcal{X^+}}  \frac{1}{k}\sum_{q \in \mathcal{N}_{x^+}} \log\frac{\exp \left[q^T \cdot f(x^+) / \tau\right]}{\exp\left[ q^T \cdot f(x^+) / \tau \right] + \sum_{x^- \in \mathcal{X^-}} \exp\left[q^T \cdot f(x^-) / \tau\right]}.
\end{equation}
The total loss is the sum of the instance discrimination loss $\mathcal{L}_\text{inst}$ and the nearest neighbors loss $\mathcal{L}_\text{nn}$ from Eq.~\ref{eq: nn_loss}: $\mathcal{L_\text{inst}} + \lambda\mathcal{L}_\text{nn}$. Figure~\ref{fig: topk} shows a schematic overview of our k-Nearest Neighbors based Momentum Contrast setup (kNN-MoCo). Algorithm~\ref{alg: knn_loss} contains the pseudocode (see also suppl.).

\definecolor{blue_fig}{RGB}{0,80,149}
\definecolor{green_fig}{RGB}{0,79,0}
\newcommand{\blue}[1]{\textcolor{blue_fig}{#1}}
\newcommand{\green}[1]{\textcolor{green_fig}{#1}}

\newcommand{\mathgreen}[1]{$\color{codeblue}{#1}$}
\begin{figure}[t]
\begin{minipage}{.55\textwidth}
\includegraphics[width=\linewidth]{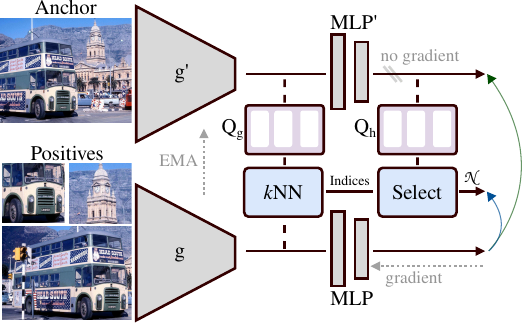}
\captionof{figure}{\textbf{kNN-MoCo setup.} $Q_g$ and $Q_h$ maintain aligned queues of backbone and output features. As before, the encoder $f$ needs to match positives with their anchor (\green{green} arrow). Additionally, we also match the positives with their $k$ nearest neighbors obtained from the queue $Q_g$ (\blue{blue} arrow).}
\label{fig: topk}
\end{minipage} 
\hfill
\begin{minipage}{.42\textwidth}
\vspace{-1.5em}
\begin{algorithm}[H]
\algcomment{\fontsize{7.2pt}{0em}\selectfont \texttt{bmm}: batch matrix mult.; \texttt{mm}: matrix mult.; \texttt{cat}: concatenate; \texttt{topk}: $k$ indices of top k elements; \texttt{CE}: cross-entropy}
\definecolor{codeblue}{rgb}{0.25,0.5,0.5}
\lstset{
  backgroundcolor=\color{white},
  basicstyle=\fontsize{7.2pt}{7.2pt}\ttfamily\selectfont,
  columns=fullflexible,
  breaklines=true,
  captionpos=b,
  commentstyle=\fontsize{7.2pt}{7.2pt}\color{codeblue},
  keywordstyle=\fontsize{7.2pt}{7.2pt},
  %frame=tb,
}
\begin{lstlisting}[language=python, escapechar=@]
@\vspace{-2.0em}@
# g, g': the (momentum-updated) backbone
# h, h': the (momentum-updated) proj. head
# x: batch of anchor images # B
@\vspace{-1.2em}@
# x@\fontsize{3pt}{0pt}\color{codeblue}$^+$@: batch of positives @\hspace{1.25em}@# B@\mathgreen{\cdot}@L
# Q@\mathgreen{_g}@: queue of x@\mathgreen{_g}@'s @\hspace{3.85em}@# C@\mathgreen{_g}@xK
# Q@\mathgreen{_h}@: queue of x@\mathgreen{_h}@'s @\hspace{3.75em}@# C@\mathgreen{_h}@xK

x@$_g$@,x@\fontsize{3pt}{0em}\selectfont$^+_g$@= g@'@(x),g(x@\fontsize{3pt}{0em}\selectfont$^+$@) @\hspace{.6em}@# BxC@\mathgreen{_g}@, B@\mathgreen{\cdot}@LxC@\mathgreen{_g}@
x@$_h$@,x@\fontsize{3pt}{0em}\selectfont$^+_h$@= h@'@(x@$_g$@),h(x@\fontsize{3pt}{0em}\selectfont$^+_g$@) @\hspace{-.0em}@# BxC@\mathgreen{_h}@, B@\mathgreen{\cdot}@LxC@\mathgreen{_h}@
x@$_g$@,x@$_h$@= x@$_g$@.detach(), x@$_h$@.detach()

l@$_{\text{pos}}$@= bmm(x@\fontsize{3pt}{0em}\selectfont$^+_h$@.view(B,L,C@$_h$@),x@$_h$@.view(B,C@$_h$@,1))
l@$_{\text{neg}}$@= mm(x@\fontsize{3pt}{0em}\selectfont$^+_h$@,Q@$_h$@).view(B,L,K)
logits = cat([l@$_{\text{pos}}$@, l@$_{\text{neg}}$@],dim=2) # BxLx(K+1)
indices = topk(mm(x@\fontsize{3pt}{0em}\selectfont$^+_g$@,Q@$_g$@).view(B,L,K),dim=2) @\hspace{1.95em}@

# loss: (1) sharpen with t, (2) apply CE
loss@$_{\text{inst}}$@ = CE(logits/t, zeros((B,L))
loss@$_{\text{nn}}$@ = multi_label_CE(l@$_{\text{neg}}$@/t, indices)
loss@$_{\text{tot}}$@ = loss@$_{\text{inst}}$@ + lambda * loss@$_{\text{nn}}$@

\end{lstlisting}
\vspace{-0.1in}
\captionof{algorithm}{Pseudocode for kNN-MoCo}
\label{alg: knn_loss}
\end{algorithm}
\end{minipage}
\end{figure}

\begin{table}[h]
    \vspace{-.8em}
    \caption{Ablation study of the number of neighbors $k$ and weight $\lambda$ for a linear classifier on VOC. $\lambda=0$ represents the multi-crop model from Section~\ref{subsec: multi_crop}. Models are trained for 200 epochs on COCO.}
    \label{tab: ablation_knn}
    \begin{minipage}{0.50\textwidth}
    \begin{center}
    \tablestyle{2pt}{1.1}
    \begin{tabular}{c |x{25} x{25} x{25} x{25} x{25}}
    $k$ & 1 & 5 & 10 & 20 & 50 \\
    \shline
    mAP (\%) & 83.6 &  84.0 & 84.1 & 84.2 & 84.3 \\
    \end{tabular}
    \end{center}
    \end{minipage}
    \hfill
    \begin{minipage}{0.50\textwidth}
    \begin{center}
    \tablestyle{2pt}{1.1}
    \begin{tabular}{c |x{25} x{25} x{25} x{25} x{25}}
    $\lambda$ & 0.0 & 0.1 & 0.2 & 0.4 & 0.8 \\
    \shline
    mAP (\%) & 82.8 &  83.8 & 84.1 & 84.2 & 80.7 \\
    \end{tabular}
    \end{center}
    \end{minipage}
\vspace{-1.5em}
\end{table}

\begin{table}[b]
\tablestyle{2pt}{1.1}
\caption{\textbf{State-of-the-art comparison.} MoCo and DenseCL are trained for 800 epochs on COCO. VirTex is trained on COCO captions~\cite{chen2015microsoft}. MoCo is trained while imposing various additional invariances.\vspace{0.5em}}
\label{tab: sota}
\begin{tabular}{ry{55}|x{35}|x{35}|x{35}|x{35}|x{35}|x{35}|x{38}|x{38} c}
\toprule
\multicolumn{2}{c|}{} &
\multicolumn{3}{c|}{\fontsize{7.5pt}{1em}\selectfont \textbf{Semantic seg. } (mIoU)} & 
\multicolumn{3}{c|}{\fontsize{7.5pt}{1em}\selectfont \textbf{Classification} (mAP / Acc. / Acc.)} &
\fontsize{7.5pt}{1em}\selectfont \textbf{Vid. Seg.} ($\mathcal{J}\&\mathcal{F}$) & 
\fontsize{7.5pt}{1em}\selectfont \textbf{Depth} (rmse) & 
\\
\multicolumn{2}{l|}{{\fontsize{7.5pt}{1em}\selectfont \textbf{Method}}} & 
{\fontsize{7.5pt}{1em}\selectfont \textbf{VOC}} &
{\fontsize{7.5pt}{1em}\selectfont \textbf{Cityscapes}} & 
{\fontsize{7.5pt}{1em}\selectfont \textbf{NYUD}} &
{\fontsize{7.5pt}{1em}\selectfont \textbf{VOC}} &
{\fontsize{7.5pt}{1em}\selectfont \textbf{ImageNet}} &
{\fontsize{7.5pt}{1em}\selectfont \textbf{Places}} & 
{\fontsize{7.5pt}{1em}\selectfont \textbf{DAVIS}} &
{\fontsize{7.5pt}{1em}\selectfont \textbf{NYUD}} &
\\
\hline
\multicolumn{2}{l|}{\demph{Rand. init.}}
& \demph{39.2} % VOC sem
& \demph{65.0} % city 
& \demph{24.4}  % nyu
& \demph{-}      % VOC 
& \demph{-}      % IN
& \demph{-}      % places
& \demph{40.8} % davis
& \demph{0.867}
& \\
\multicolumn{2}{l|}{DenseCL~\cite{wang2020dense}}
& 73.2  % VOC sem
& \textbf{73.5}  % city mask
& \textbf{42.1}  % nyu
& 83.5  % VOC svm
& 49.9  % IN
& 45.8  % places
& 61.8
& 0.589
& \\
\multicolumn{2}{l|}{VirTex~\cite{desai2020virtex}}
& 72.7
& 72.5
& 40.3
& \textbf{87.4}
& 53.8
& 40.8
& 61.3
& 0.613
& \\
\hline
\multicolumn{2}{l|}{MoCo}
& 71.1  % VOC sem
& 71.3  % city mask
& 40.0 % nyu
& 81.0  % VOC svm
& 49.8  % IN
& 44.7  % places
& 63.3 % davis
& 0.606
& \\
\multicolumn{2}{l|}{+ CC}
& 72.2  % VOC sem
& 71.6  % city mask
& 40.4  % nyu
& 84.0  % VOC svm
& 54.6  % IN
& 46.1  % places
& 65.5 % davis
& 0.595
& \\
\multicolumn{2}{l|}{+ CC + A$^+$}
& 72.7 % VOC sem
& 71.8 % city mask
& 40.7 % nyu
& 85.0
& 56.0  % IN
& 47.0  % places
& 65.7 % davis
& 0.590
& \\
\multicolumn{2}{l|}{+ CC + A$^+$ + kNN}
& \textbf{73.5} 
& 72.3
& 41.3
& 85.9
& \textbf{56.1}
& \textbf{48.6}
& \textbf{66.2}
& \textbf{0.580}
& \\
\bottomrule
\multicolumn{11}{c}{\fontsize{7.5pt}{1em}\selectfont CC: Constrained multi-crop (Sec.~\ref{subsec: multi_crop}), A$^+$: Stronger augmentations (Sec.~\ref{subsec: auto_augment}), kNN: nearest neighbors (Sec.~\ref{subsec: neighbors}).}\\
\end{tabular}
\end{table}

\paragraph{Results \& Discussion.} Table~\ref{tab: ablation_knn} contains an ablation study of the number of neighbors $k$ and the weight $\lambda$. The performance remains stable for a large range of neighbors $k$ ($\lambda$ is fixed at 0.4). However, increasing the number of neighbors $k$ positively impacts the accuracy. We use $k=20$ for the remainder of our experiments. Further, the representation quality degrades when using a large weight (e.g. $\lambda=0.8$). This shows the importance of using the instance discrimination task as our main objective. Also, not shown in the table, we found that it is important to mine the neighbors using the features before the projection head (84.2\% vs 82.8\%). Finally, Table~\ref{tab: overview_invariances} shows improved results on all tasks when combining the nearest neighbors loss with our other modifications. In conclusion, we have successfully explored the data manifold to learn additional invariances. The proposed implementation can be seen as a simple alternative to clustering-based methods~\cite{asano20self,caron2020unsupervised,li2020prototypical,wang2020unsupervised}.

\subsection{Discussion}
\label{subsec: discussion}

We retrain our final model for 800 epochs on COCO and compare with two other methods, i.e. DenseCL~\cite{wang2020dense} and Virtex~\cite{desai2020virtex}. We draw the following conclusions from the results in Table~\ref{tab: sota}. First, our model improves over the MoCo baseline. The proposed modifications force the model to learn useful features that can not be learned through standard data augmentations, even when increasing the training time. Second, our representations outperform other works on several downstream tasks. These frameworks used more advanced schemes~\cite{wang2020dense} or caption annotations~\cite{desai2020virtex}. Interestingly, DenseCL reports better results for the segmentation tasks on Cityscapes and NYUD, but performs worse on other tasks. In contrast, the performance of our representations is better balanced across tasks. We conclude that generic pretraining is still an unsolved problem. 
\section{Related Work}
\label{sec: related_work}
\paragraph{Contrastive learning.} The idea of contrastive learning~\cite{gutmann2010noise,oord2018representation} is to attract positive sample pairs and repel negative sample pairs. Self-supervised methods~\cite{chen2020simple,grill2020bootstrap,he2019momentum,misra2020self,tian2019contrastive,tian2020infomin,wu2018unsupervised,ye2019unsupervised,chen2020exploring} have used the contrastive loss to learn visual representations from unlabeled images. Augmentations of the same image are used as positives, while other images are considered as negatives. 

A number of extensions were proposed to boost the performance. For example, a group of works~\cite{wang2020dense,pinheiro2020unsupervised,van2021unsupervised,henaff2021efficient} applied the contrastive loss at the pixel-level to learn dense representations. Others improved the representations for object recognition tasks by re-identifying patches~\cite{yang2021instance} or by maximizing the similarity of corresponding image regions in the intermediate network layers~\cite{xiao2021region}. Finally, Selvaraju~\emph{et~al.}~\cite{selvaraju2020casting} employed an attention mechanism to improve the visual grounding abilities of the model. In contrast to these works, we do not employ a more advanced pretext task to learn spatially structured representations. Instead, we adopt a standard framework~\cite{he2019momentum} and find that the learned representations exhibit similar properties when modifying the cropping strategy. Further, we expect that our findings can be relevant for other contrastive learning frameworks too. 

\paragraph{Clustering.} Several works combined clustering with self-labeling~\cite{caron2018deep,asano20self} or contrastive learning~\cite{caron2020unsupervised,li2020prototypical,wang2020unsupervised} to learn representations in a self-supervised way. Similar to the nearest neighbors loss (Eq.~\ref{eq: nn_loss}), these frameworks explore the data manifold to learn invariances. Differently, we avoid the use of a clustering criterion like K-Means by computing nearest neighbors on-the-fly w.r.t. a memory bank. A few other works~\cite{huang2019unsupervised,van2020scan} also used nearest neighbors as positive pairs in an auxiliary loss. However, the neighbors had to be computed off-line at fixed intervals during training. Concurrent to our work, Dwibedi~\emph{et~al.}~\cite{dwibedi2021little} adopted nearest neighbors from a memory bank under the BYOL~\cite{grill2020bootstrap} framework. The authors focus on image classification datasets. In conclusion, we propose a simple, yet effective alternative to existing clustering methods. 

\paragraph{Other.} Contrastive SSL has been the subject of several recent surveys~\cite{purushwalkam2020demystifying,zhao2020makes,ericsson2020well}. We list the most relevant ones. Similar to our work, Zhao~\emph{et~al.}~\cite{zhao2020makes} pretrain on multiple datasets. They investigate what information is retained under the transfer learning setup, which differs from the focus of this paper. Purushwalkam and Gupta~\cite{purushwalkam2020demystifying} study the influence of the object-centric dataset bias, but their experimental scope is rather limited, and their conclusions diverge from the ones in this work. Ericsson~\emph{et~al.}~\cite{ericsson2020well} compare several ImageNet pretrained models under the transfer learning setup. In conclusion, we believe our study can complement these works.
\section{Conclusion}
\label{sec: conclusion}
In this paper, we showed that we can find a generic set of augmentations/invariances that allows us to learn effective representations across different types of datasets (i.e., scene-centric, non-uniform, domain-specific). We provide empirical evidence to support this claim. First, in Section~\ref{sec: contrastive_wild}, we show that the standard SimCLR augmentations can be applied across several datasets. Then, Section~\ref{sec: invariances} studies the use of additional invariances to improve the results for a generic dataset (i.e., MS-COCO). In this way, we reduce the need for dataset or domain-specific expertise to learn useful representations in a self-supervised way. Instead, the results show that simple contrastive frameworks apply to a wide range of datasets. We believe this is an encouraging result. Finally, our overall conclusion differs from a few recent works~\cite{purushwalkam2020demystifying,selvaraju2020casting}, which investigated the use of a more advanced pretext task or video to learn visual representations.

Our paper also yields a few interesting follow-up questions. \textbf{Modalities.} Can we reach similar conclusions for other modalities like video, text, audio, etc.? \textbf{Invariances.} What other invariances can be applied? How do we bias the representations to focus more on texture, shape or other specific properties? \textbf{Compositionality.} Can we combine different datasets to learn better representations?
\section*{Broader Impact}
The goal of this work is to study and improve contrastive self-supervised methods for learning visual representations. Self-supervised learning aims to learn useful representations without relying on human annotations. Our analysis indicates that existing methods can be applied to a large variety of datasets. This observation could benefit applications where annotated data is scarce, like medical imaging, or where large amounts of unlabeled data are readily available, like autonomous driving. Our work also improves the learned representations, and thus benefits many downstream tasks like semantic segmentation, classification, etc. These tasks are of relevance to many applications. At this point, it is hard to assess all possible societal implications of this work. After all, the advantages or disadvantages of new applications using the studied technology will depend on the intentions of the users or inventors. 
\paragraph{Acknowledgment.} The authors thankfully acknowledge support by Toyota via the TRACE project and MACCHINA (KU Leuven, C14/18/065). This work is also sponsored by the Flemish Government under the Flemish AI programme.
% for arxiv version uncomment
\newpage
% Define new colors
\definecolor{Highlight}{HTML}{39b54a}  % green
\definecolor{green}{HTML}{39b54a} % more transparent
\definecolor{red}{HTML}{cb4335} % red

\setcounter{section}{0}
\renewcommand\thesection{\Alph{section}}
\setcounter{figure}{0}
\setcounter{table}{0}
\renewcommand{\thefigure}{S\arabic{figure}}
\renewcommand{\thetable}{S\arabic{table}}

\begin{flushleft}
\Large{\textbf{Appendix}}
\end{flushleft}
%%%%%% Implementation Details
\section{Implementation Details}
This section provides all necessary details to reproduce the results from the main paper. The code will be made available upon acceptance. All pretraining experiments were run with 2 $\times$ 32GB V100 GPUs. 
\subsection{Pretraining} 
\paragraph{Datasets.} Four different datasets are used for pretraining. These include MS-COCO~\cite{lin2014microsoft}, ImageNet~\cite{deng2009imagenet}, OpenImages~\cite{kuznetsova2020open} and BDD100K~\cite{yu2020bdd100k}. All datasets are publicly available and free to use for research purposes. Images from the official train splits are used for pretraining. 
\begin{wrapfigure}[14]{r}{0.36\textwidth}
    \centering
    \includegraphics[width=1.0\linewidth]{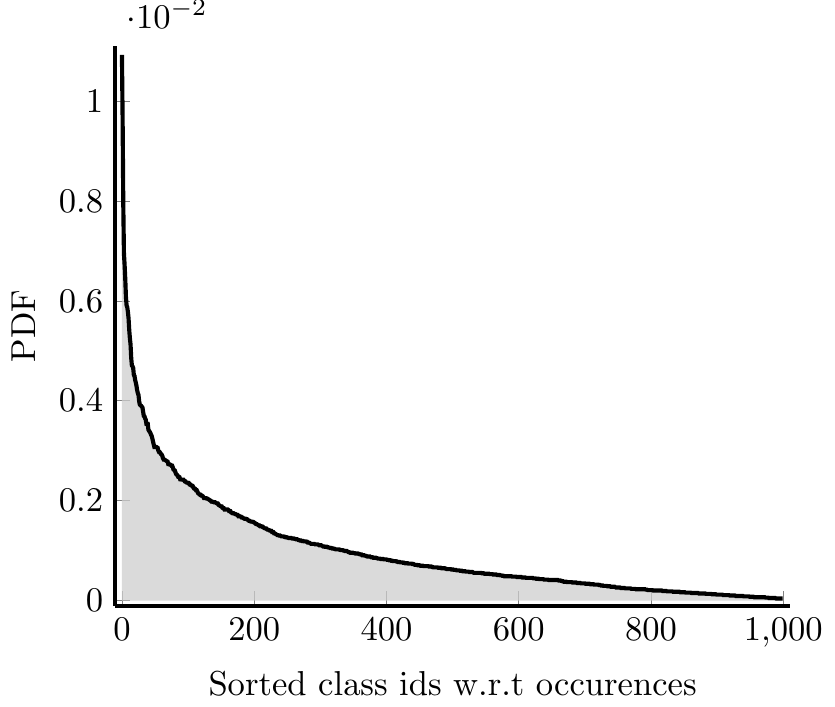}
    \captionof{figure}{Class distribution of the IN-118K-LT dataset.}
    \label{fig: class_distribution}
\end{wrapfigure}
We construct the IN-118K dataset by sampling a uniform subset from the ImageNet train split. Differently, the IN-118K-LT is a long-tailed version of the ImageNet train split. The classes follow the Pareto distribution with power value $\alpha=6$ (see Figure~\ref{fig: class_distribution}). OI-118K is obtained by randomly sampling 118K images from the OpenImages-v4 train split. Note that all datasets are constructed to be of similar size, which facilitates an apples-to-apples comparison. 

\paragraph{MoCo.} The official implementation of MoCo-v2~\cite{chen2020improved} is used to run our experiments. The model consists of a ResNet-50 backbone with MLP head. We do not include batchnorm in the head. The models in Section 3 and 4 are pretrained for 400 and 200 epochs respectively. The initial learning rate is set to 0.3. Other hyperparameters use the default values. 

\paragraph{Multi-Crop.} We modify the MoCo framework to support multi-crop training. In particular, $N$ additional smaller crops are used as positives for the anchor. Note that the crops are not processed by the momentum-encoder and do not appear as negatives in the memory bank. The crops are generated with the \texttt{RandomResizedCrop} transform from PyTorch~\cite{pytorch}. The scale is selected similar to~\cite{caron2020unsupervised}. The two large crops - the anchor and the positive - are obtained with the scale randomly selected between $[0.2,1.0]$ and output size $160\times160$ pixels. Differently, we randomly sample the scale between $[0.05,0.14]$ and use output size $96\times96$ pixels for the $N$ smaller crops. The \texttt{constrained multi-crop} is implemented by checking whether the smaller crops overlap with the anchor crop. The smaller crops are resampled if less than 20\% of their area overlaps with the anchor crop. 

\paragraph{Stronger Augmentations.} The dataloader is modified to support stronger augmentations. As before, we first apply the \texttt{RandomResizedCrop} transform to the input image. This is followed by either (1) the remaining default transformations or (2) AutoAugment~\cite{cubuk2019autoaugment}. For AutoAugment, we use the official PyTorch implementation. The augmentation policy was obtained from ImageNet~\cite{deng2009imagenet}. Since the use of stronger augmentations only impacts the input data, the rest of the setup is kept unchanged. Note that we only apply the stronger augmentations to the positives. The anchors are augmented using the default transformations. 

\paragraph{Nearest Neighbors.} The forward pass of the model is modified to incorporate the auxiliary loss. Other settings are kept. A PyTorch-like implementation of our algorithm can be found in Section~\ref{app: pseudocode}. The auxiliary loss uses weight $\lambda=0.4$ and $20$ nearest neighbors. 

\subsection{Linear classification} 
We train a linear classifier on top of the pretrained ResNet-50 backbone. The weights and batchnorm statistics of the backbone are frozen. The top-1 accuracy metric is reported for CIFAR-10~\cite{krizhevsky2009learning}, Stanford Cars~\cite{krause2013collecting}, Food-101~\cite{bossard2014food}, Pets~\cite{parkhi2012cats}, Places~\cite{zhou2017places}, SUN397~\cite{xiao2010sun} and ImageNet~\cite{deng2009imagenet}. We report the mAP metric on PASCAL VOC 2007~\cite{everingham2010pascal}. SUN397 uses the first train/test split, while other datasets use the regular train/val/test splits. All datasets are publicly available and free to use for research purposes. 

On ImageNet and Places, we adopt the evaluation protocol from He~\emph{et~al.}~\cite{he2019momentum}. The training follows the typical ResNet example in PyTorch~\cite{pytorch}. Standard data augmentations are used. The model is trained for 100 epochs using SGD with momentum $0.9$ and initial learning rate $30.0$. The linear layer is regularized with weight decay $0.0001$. The learning rate is decayed with a factor 10 after 60 and 80 epochs.

The other datasets are evaluated following the protocol from Ericsson~\emph{et~al.}~\cite{ericsson2020well}. We fit a multinomial logistic regression model to the features extracted from the backbone. The implementation from \texttt{sklearn} is used. We do not apply any augmentations, and resize all images to $224\times224$ pixels using the \texttt{Resize} transform in PyTorch. The L2 regularisation constant is selected on the validation set using 45 logarithmically spaced values between 1e-5 and 1e5. We use the L-BFGS optimizer and softmax cross-entropy objective. On PASCAL VOC, we need to solve a multi-class classification problem. In this case, we fit a binary classifier for each class.

\subsection{Segmentation} 
The representations are finetuned end-to-end for the semantic segmentation task on Cityscapes~\cite{cordts2016cityscapes}, PASCAL VOC~\cite{everingham2010pascal} and NYUD~\cite{silberman2012indoor}. All datasets are publicly available and free to use for research purposes. On Cityscapes, we train on the \texttt{train\_fine} set (2975 images) and evaluate on the \texttt{val} set. For PASCAL VOC, training is performed on the \texttt{train\_aug2012} set (10582 images) and evaluation on the \texttt{val2012} set. Finally, the \texttt{train} set (795 images) and \texttt{val} set are used for training and evaluation on NYUD. We adopt the mean intersection over union (mIoU) as evaluation metric. The training images are augmented with random scaling (ratio between 0.5 and 2.0) and horizontal flipping. 

\paragraph{PASCAL VOC.} We follow the evaluation protocol from~\cite{he2019momentum} and use an FCN-based~\cite{long2015fully} model. The $3\times3$ convolutions in the final ResNet-block have dilation 2 and stride 1. The backbone is followed by two additional $3\times3$ convolutions of 256 channels with BN and ReLU, and then a $1\times1$ convolution to obtain pixel-wise predictions. We use dilation rate 6 in the extra $3\times3$ convolutions. The model is trained for 45 epochs using batches of size 16. We use SGD with momentum 0.9 and initial learning rate 0.003. The learning rate is multiplied with $0.1$ after 30 and 40 epochs. Weight decay regularization is used of 0.0001. We  use random crops of size $513\times513$ pixels during training. 

\paragraph{Cityscapes.} The setup from PASCAL VOC is reused. The dilation rate of the $3\times3$ convolutions in the head is set to 1. This improves the results on Cityscapes, since the model can better capture small objects and classes with a thin structure like poles. The model is trained for 150 epochs using batches of size 16. We use the Adam optimizer with initial learning rate 0.0001 and a poly learning rate schedule. Weight decay regularization is used of 0.0001. Training uses random crops of size $768\times768$ pixels.

\paragraph{NYUD.} The publicly available code from~\cite{vandenhende2021multi} is used to evaluate the representations on NYUD. Again, the $3\times3$ convolutions in the final ResNet-block have dilation 2 and stride 1. The backbone is followed by a DeepLab-v3 decoder~\cite{chen2018encoder}. The model is trained for 100 epochs using batches of size 8. We use the Adam optimizer with initial learning rate 0.0001 and a poly learning rate schedule. Weight decay regularization is used of 0.0001. We use random crops of size $512\times512$ pixels during training. 

\subsection{Object Detection} 
The representations are evaluated for the \texttt{VOC 2007} object detection task~\cite{everingham2010pascal}. The default metrics are reported on the~\texttt{2007~test} split. We adopt the evaluation protocol from He~\emph{et~al.}~\cite{he2019momentum}. The code uses the \texttt{Detectron2} framework~\cite{wu2019detectron2}. The detector is Faster R-CNN~\cite{ren2015faster} with R50-dilated-C5 backbone. All layers are finetuned end-to-end. The image scale is [480, 800] pixels during training and 800 at inference. 

\subsection{Depth Estimation} 
We evaluate the representations for the depth estimation task on NYUD~\cite{silberman2012indoor}. The model and training setup are the same as for the semantic segmentation task on NYUD. We adopt the mean squared error objective function to train the model. The results report the root mean squared error (rmse) on the validation set.

\subsection{Video Instance Segmentation}
The DAVIS-2017 video instance segmentation benchmark~\cite{davis2017} is used. The dataset is publicly available and free to use for research purposes. The mean region similarity $\mathcal{J}_m$ and mean contour-based accuracy $\mathcal{F}_m$ are reported on the \texttt{2017 val} split. We adopt the evaluation protocol from Jabri~\emph{et~al.}~\cite{jabri2020walk}. The labels are propagated using nearest neighbors between consecutive frames. Note that we do not apply any finetuning for the task. 

\subsection{Semantic Segment Retrieval}
\label{app: segments}
Finally, we evaluate the representations for the semantic segment retrieval task on PASCAL VOC~\cite{everingham2010pascal}. The train and val splits are the same as for the semantic segmentation task. The model is a ResNet-50. The $3\times3$ convolutions in the final ResNet-block use dilation 2 and stride 1. The average pooling layer is removed to maintain the spatial structure of the output. The input images are rescaled to $512\times512$ pixels and the output is of size $64\times64$ For each image, we cluster the spatial features with K-means (K=15). This results in K regions per image. For each region, we compute a region descriptor by averaging the features of all pixels within the region. The validation regions are assigned a label by obtaining the nearest neighbor from the train set. Note that we do not use a weighted k-NN algorithm. The predictions are rescaled to match the original input resolution for both evaluation and visualization. The mean intersection over union (mIoU) is reported on the validation set.

%%%% Pseudocode
\section{Pseudocode}
\label{app: pseudocode}
Algorithm~\ref{alg: forward_pass} provides the pseudocode of our framework. The code supports the \texttt{constrained multi-crop}, stronger augmentations and the nearest neighbors auxiliary loss. The implementation builds on top of MoCo. However, the ideas in this paper could be combined with other frameworks too. We highlight the most important differences with MoCo. In order to mine nearest neighbors that are visually similar, the representations should already model semantically meaningful properties. However, at the start of training, the model is initialized randomly. In this case, we can not learn any useful invariances from grouping nearest neighbors. We resolve this problem by only applying the auxiliary loss after a fixed number of epochs (5). 

We found an alternative implementation which allows to adopt the auxiliary loss from the start of training. We can update the memorybank (final line in~Algorithm~\ref{alg: forward_pass}) before computing the loss. In this case, there is a mismatch between the logits and the memory bank for the most recent batch~\footnote{The memory bank is implemented as an array of fixed size. We replace the oldest batch when adding a new batch using a pointer.}. This means we will inject random noise when the nearest neighbors map to the most recently used batch in the memory bank. We observe that this behavior mostly occurs early on during training, as the memory bank is initialized randomly. Over time, this effect is reduced, and we end up using a loss that is similar to what we had before. We report similar results for both strategies. 

\begin{algorithm}[H]
\newcommand{\hlp}{\makebox[0pt][l]{\color{Mulberry!15}\rule[-3pt]{0.72\linewidth}{9pt}}}
\newcommand{\hlb}{\makebox[0pt][l]{\color{NavyBlue!15}\rule[-3pt]{0.72\linewidth}{9pt}}}
\algcomment{\fontsize{7.2pt}{0em}\selectfont \texttt{bmm}: batch matrix mult.; \texttt{mm}: matrix mult.; \texttt{cat}: concatenate; \texttt{topk}: indices of top-k largest elements; \texttt{CE}: cross-entropy loss \\  Important differences with MoCo are highlighted.}
\definecolor{codeblue}{rgb}{0.25,0.5,0.5}
\lstset{
  backgroundcolor=\color{white},
  basicstyle=\fontsize{7.2pt}{7.2pt}\ttfamily\selectfont,
  columns=fullflexible,
  breaklines=true,
  captionpos=b,
  commentstyle=\fontsize{7.2pt}{7.2pt}\color{codeblue},
  keywordstyle=\fontsize{7.2pt}{7.2pt},
  %frame=tb,
}
\begin{lstlisting}[language=python, escapechar=@]
@\vspace{-2.0em}@
# g, g_m: the backbone g and momentum updated backbone g_m
# h, h_m: the projection head h and momentum updated projection head h_m (MLP)
# anchors: batch of anchor images (B x 3 x 224 x 224)
# positives: batch of global positive views (B x 3 x 224 x 224)
# positives_small: batch of N local positive views (B*N x 3 x 96 x 96)
# queue: dictionary as a queue (C x K)
# queue_g: dictionary as a queue (C_g x K)
# m: momentum
# t: temperature
# k: number of nearest neighbors
# lambda: weight in range [0, 1]

g_m.params = g.params # initialize momentum updated backbone
h_m.params = h.params # initialize momentum updated head

for batch in loader: # load a minibatch with B samples
   
    # randomly augment batch
    anchors @~~@= aug(batch)
    positives = aug(batch)
    @\hlb@positives_small = aug_small(batch, anchors) # constrained multi-crop @~~~~~~~~~\textcolor{red}{Sec.~4.1}@
    
    # forward pass
    anchors_g, positives_g = g_m(anchors), g(positives) @\hspace{4em}@# B x C_g
    anchors, positives = h_m(anchors_g), h(positives_g) @\hspace{4em}@# B x C
    @\hlb@positives_small_g = g(positives_small) @\hspace{10.5em}@# B*N x C_g 
    @\hlb@positives_small = h(positives_small_g) @\hspace{10.5em}@# B*N x C 
    
    # concatenate positive views
    @\hlb@positives_g = cat([positives_g, positives_small_g], dim=0) @\hspace{0.5em}@# B*(N+1) x C_g
    @\hlb@positives = cat([positives, positives_small], dim=0) @\hspace{3.5em}@# B*(N+1) x C

    # compute logits
    l_pos = bmm(positives.view(B,N+1,C), anchors.view(B,C,1))@\hspace{1.5em}@# B x N+1 x 1
    l_neg = mm(positives, queue.view(C,K)).view(B,N+1,K) @\hspace{3.5em}@# B x N+1 x K
    logits = cat([l_pos, l_neg], dim=2) @\hspace{12em}@# B x N+1 x K+1
    
    # determine indices of the nearest neighbors
    @\hlp@indices = topk(mm(positives_g, queue_g).view(B,N+1,K),dim=2)# B x N+1 x k @~~~~\textcolor{red}{Sec.~4.3}@

    # loss: (1) sharpen with temperature t, (2) apply cross-entropy loss
    loss_inst = CE(logits/t, zeros((B,N+1))
    @\hlp@loss_nn = multi_label_CE(l_neg/t, indices)
    @\hlp@loss = loss_inst + lambda * loss_nn 
    
    # SGD update: g, h
    loss.backward()
    update(g.params, h.params)
    
    # momentum update: g_m, h_m
    g_m.params = m * g_m.params + (1-m) * g.params
    h_m.params = m * h_m.params + (1-m) * h.params

    # update dictionaries: queue & queue_g
    enqueue_dequeue(queue, anchors) 
    @\hlp@enqueue_dequeue(queue_g, anchors_g) 
\end{lstlisting}
\vspace{-0.1in}
\captionof{algorithm}{Pseudocode of kNN-MoCo in a PyTorch-like style.}
\label{alg: forward_pass}
\end{algorithm}

%%%% Scaling
\section{Additional Pretraining Results}
\label{app: scaling}
The experiments in the main paper were mostly performed on datasets of moderate size (e.g. MS-COCO). In this section, we include additional results when pretraining on datasets of larger size. Again, we compare the use of object-centric (ImageNet) versus scene-centric (OpenImages) images for pretraining. We start from the IN-118K and OI-118K datasets defined in the main paper, and increase the size of the dataset by a factor of 2, 4 and 8. The representations are again evaluated by transferring them to multiple dense prediction tasks. 

Table~\ref{tab: scaling} shows the results. We draw the following two conclusions. First, as expected, the numbers improve when more data is used. However, the results show diminishing returns when scaling up the dataset size. He~\emph{et~al.} made a similar observation. Second, there are no significant disadvantages to pretraining on ImageNet versus OpenImages, even when using more data. This result is in line with the observations from the main paper.

\begin{table}[!t]
    \vspace{-1.5em}
    \caption{Comparison of representations trained on datasets of varying size. Results are obtained when applying end-to-end finetuning. We indicate the differences with the $1\times$ baseline.}
    \label{tab: scaling}
    \tablestyle{.8pt}{1.1}
    \centering
    \begin{tabular}{ry{50}|x{30}|x{30}|x{44}|x{44}|x{44}|x{62}|x{50} c}
    \toprule
    ~ & ~ & ~ & ~ &\multicolumn{3}{c|}{\fontsize{7.5pt}{1em}\selectfont \textbf{Semantic seg. } (mIoU)} &
    {\fontsize{7.5pt}{1em}\selectfont \textbf{Vid. seg.~ ($\mathcal{J}\&\mathcal{F})$}} & {\fontsize{7.5pt}{1em}\selectfont \textbf{Depth} (rmse)} & \\
    \multicolumn{2}{l|}{{\fontsize{7.5pt}{1em}\selectfont \textbf{Pretrain dataset}}} &
     {\fontsize{7.5pt}{1em}\selectfont \textbf{Size}} &
    {\fontsize{7.5pt}{1em}\selectfont \textbf{Factor}} &
    {\fontsize{7.5pt}{1em}\selectfont \textbf{VOC}} &
    {\fontsize{7.5pt}{1em}\selectfont \textbf{Cityscapes}} & 
    {\fontsize{7.5pt}{1em}\selectfont \textbf{NYUD}} &
    {\fontsize{7.5pt}{1em}\selectfont \textbf{DAVIS}} &  
    {\fontsize{7.5pt}{1em}\selectfont \textbf{NYUD}} & \\
    \hline
    \multicolumn{2}{l|}{IN-118K} & 118K & \multicolumn{1}{c|}{{$1\times$}} & \ressup{68.9}{} & \ressup{70.1}{} & \ressup{37.7}{} &  \ressup{63.5}{} & \ressupnyud{0.625}{} & \\
    
    \multicolumn{2}{l|}{} & 236K & \multicolumn{1}{c|}{{$2\times$}} & \reshg{72.4}{+}{3.5} & \reshg{71.1}{+}{1.0} & \reshg{39.9}{+}{2.2} & \reshg{64.9}{+}{1.4} & \reshgnyud{0.612}{-}{0.013} & \\
    
    \multicolumn{2}{l|}{} & 472K & \multicolumn{1}{c|}{{$4\times$}} & \reshg{73.5}{+}{4.6} & \reshg{71.1}{+}{1.0} & \reshg{39.9}{+}{2.2} & \reshg{64.9}{+}{1.4} & \reshgnyud{0.607}{-}{0.018} & \\
    
    \multicolumn{2}{l|}{} & 944K & \multicolumn{1}{c|}{{$8\times$}} & \reshg{75.0}{+}{6.1} & \reshg{71.7}{+}{1.6} & \reshg{40.9}{+}{3.2} & \reshg{66.1}{+}{2.6} & \reshgnyud{0.599}{-}{0.026} & \\
    
    \shline
    
    \multicolumn{2}{l|}{OI-118K} & 118K &\multicolumn{1}{c|}{{$1\times$}} & \ressup{67.9}{} & \ressup{70.9}{} & \ressup{38.4}{} &  \ressup{64.8}{} & \ressupnyud{0.609}{} & \\
    
    \multicolumn{2}{l|}{} & 236K & \multicolumn{1}{c|}{{$2\times$}} & \reshg{71.4}{+}{3.5} & \reshg{71.4}{+}{0.5} & \reshg{40.1}{+}{1.7} & \reshr{64.5}{-}{0.3} & \reshgnyud{0.609}{{\transparent{0}+}}{0.0} & \\
    
    \multicolumn{2}{l|}{} & 472K & \multicolumn{1}{c|}{{$4\times$}} & \reshg{72.6}{+}{4.7} & \reshg{71.9}{+}{1.0} & \reshg{40.9}{+}{2.5} & \reshr{64.2}{-}{0.6} & \reshgnyud{0.601}{-}{0.008} & \\
    
    \multicolumn{2}{l|}{} & 944K &\multicolumn{1}{c|}{{$8\times$}} & \reshg{73.6}{+}{5.7} & \reshg{72.2}{+}{1.3} & \reshg{41.0}{+}{2.6} & \reshg{65.0}{+}{0.2} & \reshgnyud{0.600}{-}{0.009} & \\
    
    \bottomrule
    \end{tabular}
    %\end{subtable}
\end{table}

\section{Class Activation Maps}
\begin{figure}
    \centering
    \includegraphics[width=\textwidth]{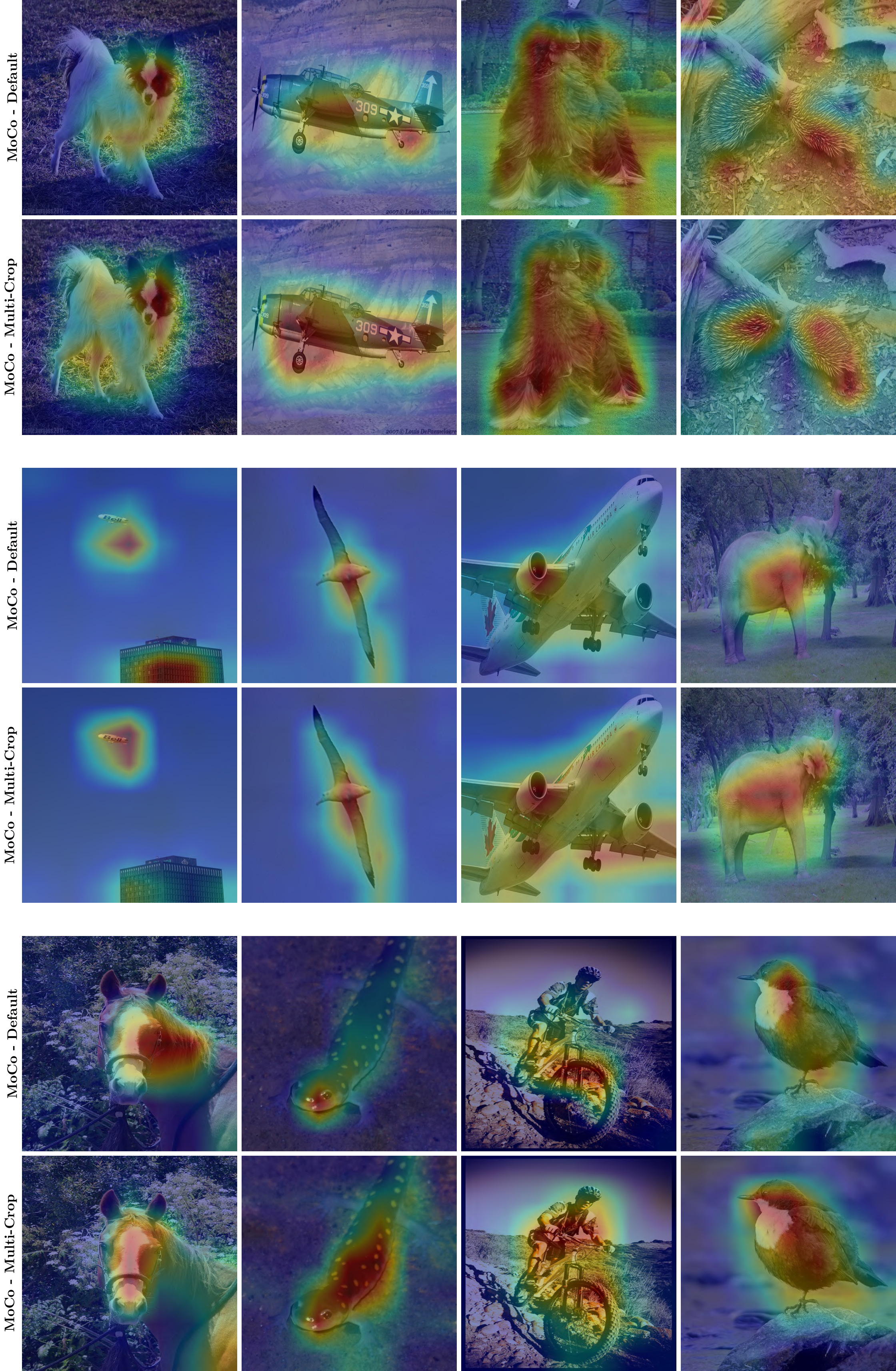}
    \caption{We show class activation maps for linear classifiers trained on top of frozen representations for ImageNet. Results are shown for MoCo trained with the two-crop (\emph{top rows}) or multi-crop (\emph{bottom rows}) transform. We trained for 200 epochs on MS-COCO.}
    \label{fig: cam}
\end{figure}

Figure~\ref{fig: cam} shows class activation maps~\cite{zhou2016learning} obtained from training linear classifiers on top of frozen representations for ImageNet. The representations were obtained by training MoCo for 200 epochs on MS-COCO. The default two-crop (top rows) or multi-crop (bottom rows) transformations were used to generate positive pairs. We observe that the multi-crop model is better at localizing the object of interest. In particular, the two-crop model often attends to only a few parts of the object of interest. Differently, the class activation maps obtained with the multi-crop model seem to segment the complete object. We believe these results could be useful for researchers working on weakly-supervised semantic segmentation methods as well. The latter group of works used class activation maps to obtain a segmentation from annotations that are easy to obtain (e.g. image-level tags~\cite{papandreou2015weakly,tang2018regularized}). 

\section{Video Instance Segmentation}
\begin{figure}
    \centering
    \includegraphics[width=\textwidth]{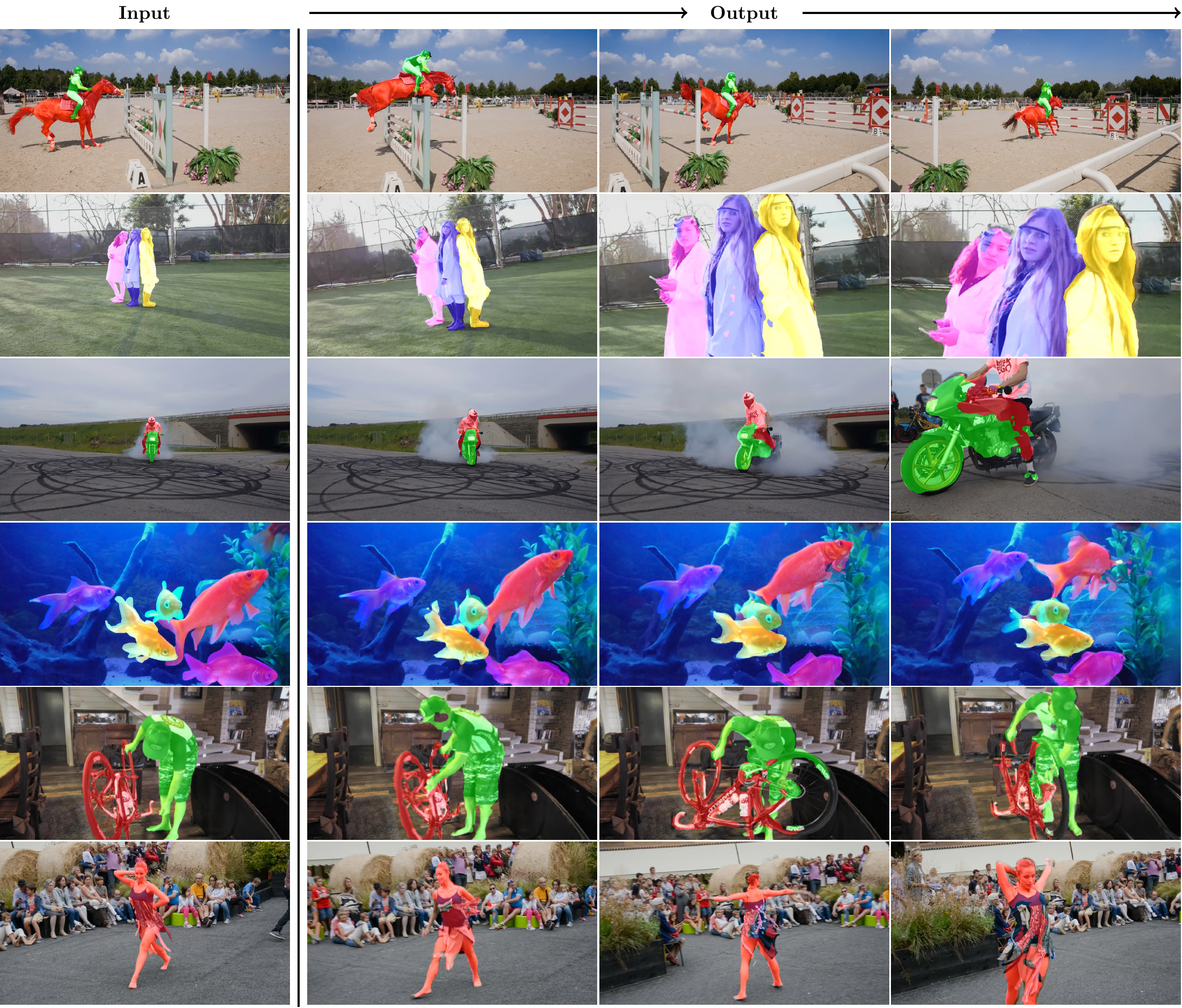}
    \caption{\textbf{DAVIS 2017 video instance segmentation.} Results are shown when training MoCo with the \texttt{multi-crop} transform for 200 epochs on MS-COCO. The labels are propagated for frozen representations via nearest neighbors.}
    \label{fig: davis_supp}
\end{figure}

Figure~\ref{fig: davis_supp} shows additional qualitative results for the video instance segmentation task on DAVIS-2017. The input consists of an annotated frame. The annotations are propagated across frames by using nearest neighbors following~\cite{jabri2020walk}. Note that the evaluation does not require any finetuning. We conclude that the representations 
can be used to model dense correspondences in videos. 

\section{Semantic Segment Retrieval}
\begin{figure}
    \centering
    \includegraphics[width=\textwidth]{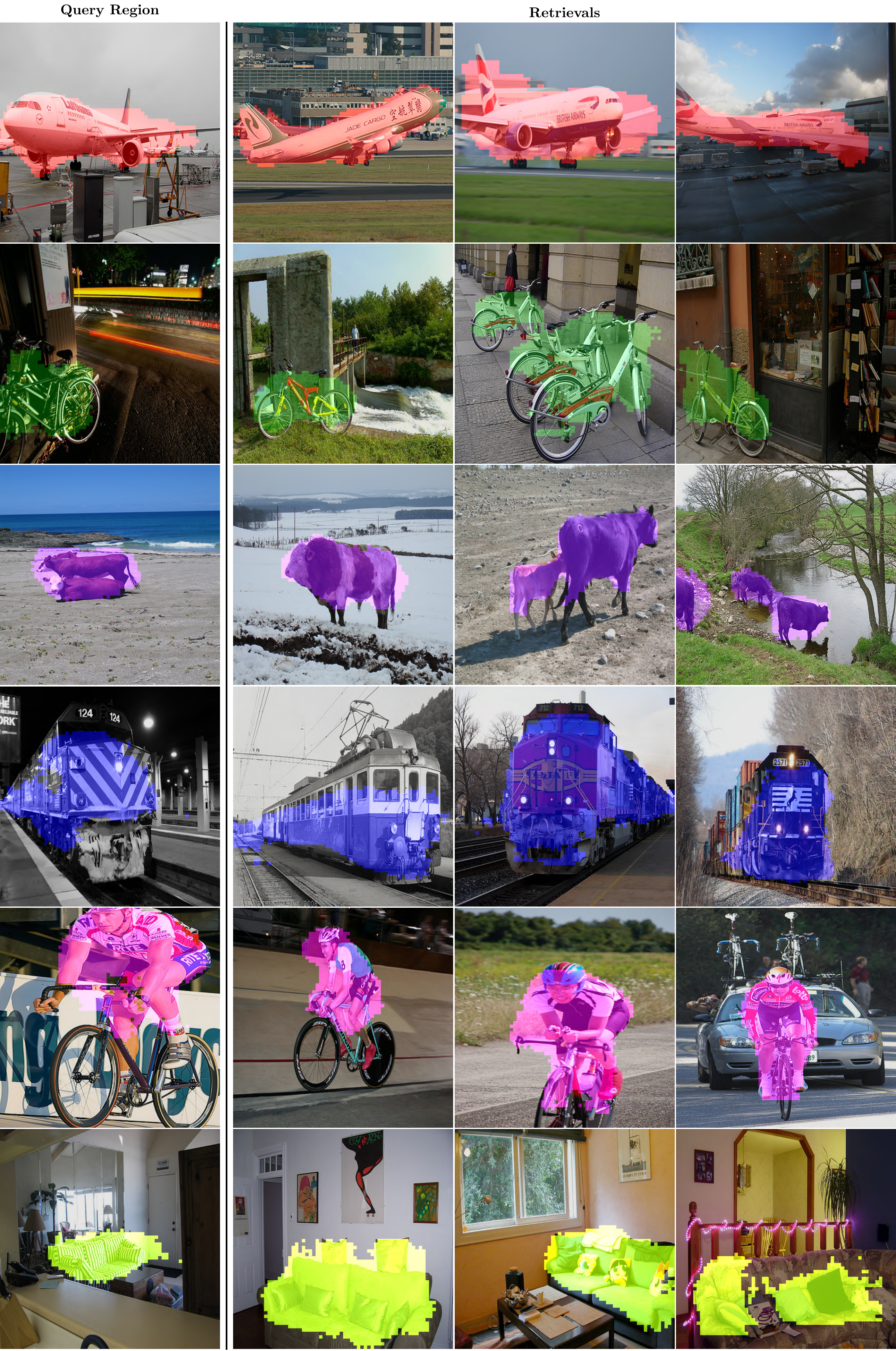}
    \caption{\textbf{PASCAL VOC semantic segment retrieval.} Results are shown when training MoCo with the \texttt{multi-crop} transform for 200 epochs on MS-COCO.}
    \label{fig: segments_supp}
\end{figure}

Figure~\ref{fig: segments_supp} shows additional qualitative results for the semantic segment retrieval task on PASCAL VOC (see also~\ref{app: segments}). The input consist of a query region obtained from applying K-Means to the input image. We compute a region descriptor by averaging the features of all the pixels within the region. Finally, we retrieve the nearest neighbors for the query. 

%%%% Limitations
\section{Limitations}
We performed extensive experiments to measure the influence of different dataset biases on the representations. The experimental setup covered the use of object-centric vs. scene-centric data, uniform vs. long-tailed class distributions and general vs. domain-specific data. Further, we explored various ways of imposing additional invariances to improve the representations. Undoubtedly, there are several components that fall outside the scope of our study. We briefly discuss some of the limitations below.

\paragraph{Other SSL methods.} In this paper, we performed our experiments using the MoCo framework. It still remains an open question whether our findings will also translate to other self-supervised methods like SimCLR~\cite{chen2020simple}, SWAV~\cite{caron2020unsupervised}, BYOL~\cite{grill2020bootstrap}, etc. 

\paragraph{Dataset size.} The experiments in the main paper were conducted on datasets of moderate size (e.g. MS-COCO contains 118K images). It remains an open question whether the same results will be observed for larger datasets. We partially addressed this by repeating a subset of our experiments on datasets of increasing size in Section~\ref{app: scaling}. However, these experiments still only considered datasets with fewer than a million images. Moreover, it becomes increasingly difficult to isolate specific properties of the data when using more samples. For example, its not obvious that a uniform dataset of 1 billion images exists. In conclusion, we believe it would be useful to further investigate the behavior of existing methods on large-scale datasets.

\paragraph{Data augmentations.} Imposing invariances to different data transformations proves crucial to learn useful representations. In this work, we used the same data augmentation strategy for all our experiments. However, different types of data could benefit from a different set of augmentations. We hypothesize that this is particularly true for domain-specific datasets. In this case, we expect that specialized data transformations, based upon domain-knowledge, could further boost the results. Further, we observe that the performance of existing methods still strongly depends on handcrafted augmentations, which can limit their applicability. We tried to partially alleviate this problem by adopting nearest neighbors - which only relies on the data manifold itself. Still, this problem could benefit from further investigation. 

\paragraph{Inductive biases.} All experiments were performed with a ResNet architecture. It would be interesting to study how architectural design choices influence the representation quality. In particular, one could study more general models like transformers~\cite{dosovitskiy2020image,vaswani2017attention} which incorporate different biases. 

\newpage
{\small
\bibliographystyle{splncs04}
\bibliography{egbib}
}

\end{document}